%% file: main.tex
\documentclass{applemlr}
\usepackage{amsmath}
\usepackage{enumerate} 
\usepackage{algorithm}
\usepackage{algpseudocode}
\usepackage{amsfonts}
\usepackage{amsthm}
\usepackage{newtxtt}
\usepackage{cleveref}
\usepackage{diagbox} 
\usepackage{colortbl}
\usepackage{amssymb}
\usepackage{xspace}
\usepackage{wrapfig}
\usepackage{adjustbox}
\usepackage{tabularx}
\usepackage{booktabs}
\usepackage{mathtools}
\usepackage{tikz}
\usepackage{paralist}

\usepackage{silence}

\definecolor{lightgreen}{RGB}{145, 204, 117}
\definecolor{lightyellow}{RGB}{250, 200, 88}
\definecolor{lightred}{RGB}{238, 102, 102}
\definecolor{lightblue}{RGB}{115, 192, 222}
\definecolor{lightorange}{RGB}{253, 208, 162} 
\newtcolorbox{promptbox}[3][Judge Prompt]{
colback=black!5!white,
arc=5pt, 
boxrule=0.5pt,
fonttitle=\bfseries,
title=#1, 
before upper={\small}, fontupper=\fontfamily{ptm}\selectfont,
colframe=#2,
label=#3,
}

\usepackage{dsfont}
\usepackage{multirow}
\usepackage{makecell}
\usepackage{enumitem}

\usepackage{xfakebold}
\input{math_commands}

\newcommand{\px}{p(y\,|\,x)}
\newcommand{\pd}{p(d\,|\,x)}
\newcommand{\qyd}{p(y\,|\,x,d)}
\newcommand{\sd}{s_{xd}}

\definecolor{findingbg}{HTML}{E8F1FA}  
\definecolor{findingframe}{HTML}{A7C4E0}

\definecolor{textgray}{HTML}{6E6E73}
\usetikzlibrary{positioning, calc}
\usetikzlibrary{decorations.pathmorphing}

\makeatletter
\patchcmd{\wrong@fontshape}{\@gobbletwo}{}{}{}
\makeatother
\numberwithin{equation}{section} 
\tcbuselibrary{minted}
\setminted[python]{
  linenos,
  breaklines,
  fontsize=\footnotesize,
  xleftmargin=2em
}

\makeatletter
\AtBeginDocument{
  \urlstyle{sf}
  
}
\makeatother

\usepackage{url}

\usepackage{hyperref}

\definecolor{light}{RGB}{125, 125, 125}
\crefname{tcb@cnt@pbox}{code}{code}
\Crefname{tcb@cnt@pbox}{Code}{Code}
\crefname{assumption}{assumption}{assumption}
\Crefname{assumption}{Assumption}{Assumptions}

\newtcolorbox[auto counter]{pbox}[2][]{
  colback=white,
  title=Code~\thetcbcounter: #2,
  #1,fonttitle=\sffamily,
  fontupper=\sffamily,
  arc=2pt,
  colframe=bgcolor,
  coltitle=fgcolor,
  colbacktitle=bgcolor,
  toptitle=0.25cm,
  bottomtitle=0.125cm
}

\makeatletter
\newcommand\applefootnote[1]{%
  \begingroup
  \renewcommand\thefootnote{}%
  \renewcommand\@makefntext[1]{\noindent##1}%
  \footnote{#1}%
  \addtocounter{footnote}{-1}%
  \endgroup
}
\makeatother

\definecolor{cverbbg}{gray}{0.90}

\usepackage{soul}

\title{CLaRa: Bridging Retrieval and Generation with Continuous Latent Reasoning}

\author{
  Jie He$^{1,2}$,
  Richard He Bai$^{1}$,
  Sinead Williamson$^{1}$,
  Jeff Z. Pan$^{2}$,
  Navdeep Jaitly$^{\dagger1}$,
  Yizhe Zhang$^{1}$
}

\affiliation{
  $^{1}$Apple \quad
  $^{2}$University of Edinburgh
}

\abstract{
Retrieval-augmented generation (RAG) enhances large language models (LLMs) with external knowledge but still suffers from long contexts and disjoint retrieval–generation optimization.
In this work, we propose CLaRa (Continuous Latent Reasoning), a unified framework that performs embedding-based compression and joint optimization in a shared continuous space.
To obtain semantically rich and retrievable compressed vectors, thereby reducing the document length fed into the generator, we introduce SCP, a key-preserving data synthesis framework based on question-answering and paraphrase supervision.
CLaRa then trains the reranker and generator end-to-end via a single language modeling loss, with gradients flowing through both modules using a differentiable top-k estimator.
Theoretically, this unified optimization aligns retrieval relevance with answer quality.
Experiments across multiple QA benchmarks show that CLaRa achieves state-of-the-art compression and reranking performance, even at a text compression rate of 16, outperforming text-based fine-tuned baselines. }

\metadata[Code]{\url{{https://github.com/apple/ml-clara}}}
\metadata[Correspondence]{\sffamily Jie He: \url{j.he@ed.ac.uk}; Yizhe Zhang: \url{yizhe_zhang@apple.com}}
\date{\sffamily\today}
\usepackage{natbib}

\begin{document}

\maketitle
\applefootnote{ \textcolor{textgray}{\sffamily $\dagger$ Work done at Apple. }}

\section{Introduction}
\textit{Retrieval-Augmented Generation} (RAG) has become a powerful paradigm for enhancing LLMs across diverse NLP tasks~\citep{10.5555/3495724.3496517,gao2024retrievalaugmentedgenerationlargelanguage,li-etal-2024-retrieval,DBLP:journals/corr/abs-2407-13193,abootorabi-etal-2025-ask}.
By incorporating external evidence, RAG mitigates key weaknesses of LLMs such as hallucination~\citep{ayala-bechard-2024-reducing} and knowledge obsolescence~\citep{lau2025readingtimelinesraganswering}.

Most RAG systems suffer from a fundamental structural issue: \textbf{retrieval and generation are optimized separately}. Retrievers select documents based on surface-level similarity, while generators produce answers without providing feedback about what information is truly needed~\citep{shi2025directretrievalaugmentedoptimizationsynergizing}. This leads to two intertwined challenges.
\textbf{(1) Efficiency.} Dense retrievers rank documents in embedding space, while generators still consume raw text, resulting in an architectural mismatch. This mismatch yields (i) inconsistent representation spaces that prevent end-to-end optimization, (ii) redundant text processing that increases inference cost~\citep{10.5555/3692070.3693515} and causes context overflow~\citep{leng2024longcontextragperformance,yue2025inference}, and (iii) duplicated encoding for both retrieval and generation. Even if gradients could flow jointly, these inefficiencies would persist due to the lack of a shared latent space.
\textbf{(2) Optimization.} Because document selection is discrete, gradients cannot flow from the generator back to the retriever~\citep{sachan2021endtoend,lin2024radit}, hindering joint training and preventing the retriever from aligning with the generator’s task objective.

\begin{figure*}[t]
\centering
\includegraphics[width=0.8\linewidth]{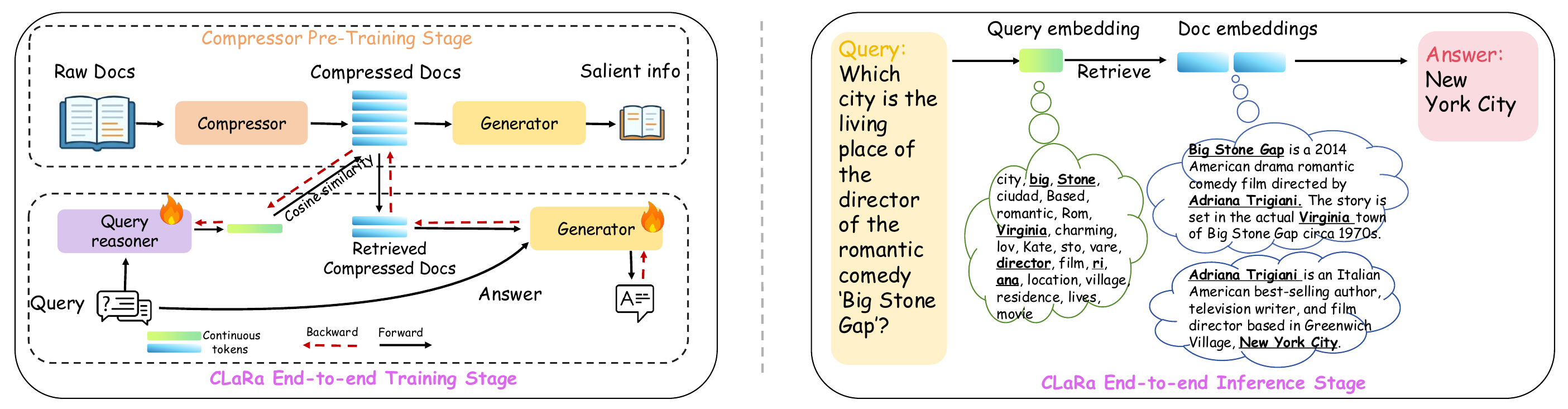}
\caption{
(a) During training, we first \textbf{pretrain the compressor} to encourage it to retain only essential information. Next, we \textbf{perform offline compression} of the documents. After that, we encode the query using the \textbf{query reasoner}, retrieve the compressed document representations for generation, and \textbf{use only the final next-token prediction loss} to jointly update both the query reasoner and the generator.
  (b) An example from the inference stage: the \textbf{\underline{tokens}} represent key clue words related to the question. When we decode the continuous query embedding, we find that it contains information not present in the original query, indicating that it has learned some of the intermediate reasoning keywords.
}
\label{fig:intro}
\vspace{-0.3cm}
\end{figure*}

\paragraph{Our Key Insight: Shared Continuous Representations.}

To address the aforementioned issues, we propose a unified framework that performs retrieval and generation over \textbf{shared continuous document representations}, as illustrated in Fig.~\ref{fig:intro}.  
Each document is encoded only once into a compact set of memory tokens, which simultaneously serve both retrieval and generation.
This design is motivated by:
\textbf{(1)} Not all tokens in textual documents carry informative content. Text can be compressed into continuous representations with high representational capacity that preserve only the most salient information, and more compact document representations can also effectively reduce the input length of the generator.  
\textbf{(2)} Moreover, continuous representations and joint optimization are inherently complementary: continuous encodings make the retrieval process differentiable, while joint training aligns the retriever and the generator within a shared semantic space optimized for reasoning.

We design a  salient-information–aware pretraining objective that pushes the model to compress documents into compact encoded representations. The generator can then answer questions by relying solely on these shorter input representations. Meanwhile, we backpropagate the generator’s next-token prediction (NTP) loss to the retriever, providing a weak supervision signal that naturally adapts retrieval to downstream generation objectives. This mechanism enables the retriever to learn which documents truly contribute to answer generation, rather than relying only on surface-level similarity. This unified design simultaneously addresses both challenges.
\textbf{Efficiency-wise}, shared encodings eliminate redundant computation, making true end-to-end optimization and inference within a unified representation space possible. Moreover, during generation, the use of shorter context lengths saves precious resources when large language models process long-form inputs.
\textbf{Optimization-wise}, continuous representations enable differentiable top-$k$ selection via Straight-Through (ST) estimation  \citep{bengio2013estimating}, allowing generator gradients to update the retriever directly through gradient descent rather than inefficient RL sampling.

To realize this vision, we present \textbf{CLaRa} (\textit{Continuous Latent Reasoning}), a joint retrieval–generation framework built on shared compressed representations.
In Stage I, we propose \textbf{SCP} (\textit{Salient Compressor Pretraining}), which enhances semantic fidelity by constructing QA pairs that emphasize salient document content beyond surface reconstruction.
In Stage II, CLaRa performs \textbf{end-to-end joint training} of the query encoder and answer generator under a unified next-token prediction loss, with \textbf{differentiable top-$k$ selection} via ST estimation.
Theoretically, we show this unified objective yields valid gradients for retriever learning without explicit labels.

We evaluate CLaRa on four single-hop and multi-hop QA benchmarks with \textit{Mistral-7B} and \textit{Phi-4B}.
Results show that SCP produces semantically rich compressed representations, and CLaRa achieves state-of-the-art retrieval and generation performance—outperforming both supervised and unsupervised baselines, and even surpassesing text-only DRO methods when our text compression ratio is 16.

\begin{figure*}[t]
    \centering
    \includegraphics[width=0.8\linewidth]{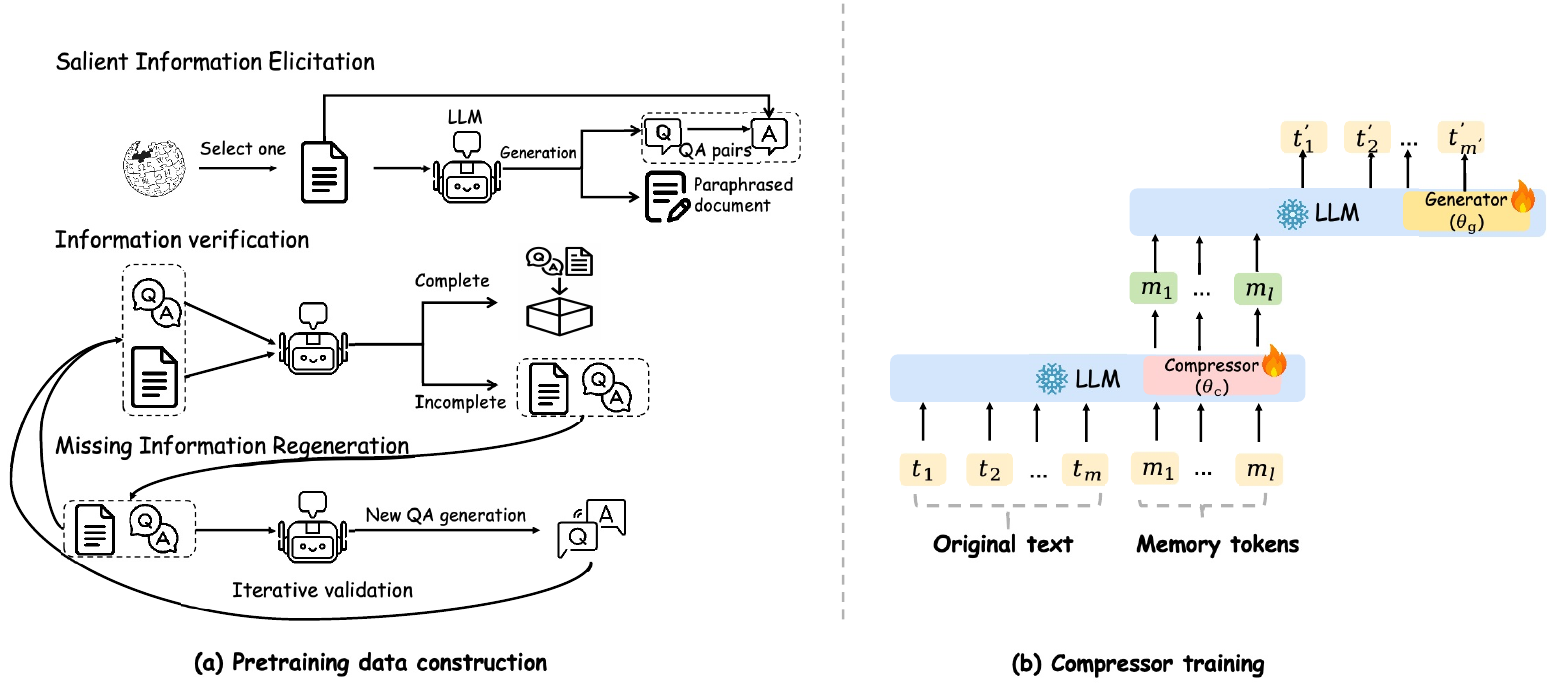}
    \caption{Overview of the SCP 
    framework, featuring (a) synthetic pretraining data construction; 
    (b) compressor training using this data.}
    \label{fig:datagen}
    \vspace{-0.3cm}
\end{figure*}

\section{SCP: Salient Compressor Pretraining}
\label{sec:method}

Previous methods \citep{louis-etal-2025-pisco, 10.5555/3737916.3741392} typically use token-level reconstruction loss to learn document representation. This risks wasting limited capacity/budget on potentially trivial token-by-token reconstruction, and the resulting representations might not ``digest'' the document in a way that retains all relevant information.  To enable the model to focus on generating semantically informative representations, we first synthesize pre-training data that highlights salient information. Based on this data, we train a compression framework, where a compressor learns to retain just the essential semantics (Fig~\ref{fig:datagen}).

\subsection{Guided Data Synthesis for Semantic Preservation}

We first construct a synthetic dataset where salient information is explicitly exposed through QA and paraphrasing. This way, the compressor later learns to identify, digest and retain the \textbf{semantic core} of the text by deeply processing the raw token-level information. 
As shown in Fig~\ref{fig:datagen} (a), our synthetic data generation pipeline consists several steps:  
(1) salient information elicitation via QA and paraphrase generation,  
(2) automatic verification of coverage, and  
(3) regeneration of missing content.

\paragraph{Salient Information Elicitation.}

Using 2M sampled Wikipedia-2021 documents~\citep{10.5555/3648699.3648950}, a locally deployed LLM (\textit{Qwen-32B}) generates three complementary supervision signals:
\begin{itemize}
\setlength{\itemsep}{0pt}
\setlength{\parsep}{0pt}
\setlength{\parskip}{0pt}
\item \textbf{Simple QA:} each pair targets a single atomic fact to encourage fine-grained factual retention, while explicitly avoiding redundancy by extracting distinct facts not covered by previous questions.  

\item \textbf{Complex QA:} each pair integrates multiple facts to promote relational reasoning, with generation guided to connect previously unlinked information and increase coverage.  

\item \textbf{Paraphrase:} paraphrased documents alter surface structure while preserving semantics, encouraging the model to learn semantic representations through a continuous information bottleneck.
\end{itemize}

QA pairs distill fact-centric supervision as they tell the model which details are essential for answering meaningful questions.
Paraphrases, in contrast, demonstrate expression, level compactness, how to rephrase the same content more efficiently.
Together, they form complementary signals: factual grounding and linguistic compactness.

\paragraph{Verification and Regeneration.}
Each document and its generated outputs (QA pairs or paraphrases) are verified by the locally deployed LLM for factual consistency and information coverage.  
When missing information is detected, the LLM reviews both the original text and existing QA pairs to generate additional ones capturing uncovered facts, iteratively up to ten rounds.  
Samples failing final coverage criteria are excluded. This iterative check ensures the model only learns from fully covered, factually faithful pairs.

\subsection{Compressor Pretraining}
\label{sec:scp}

Following  \textit{PISCO}~\citep{louis-etal-2025-pisco}, we adopt a shared base model equipped with multiple \textsc{LoRA} adapters for modular control, where each adapter corresponds to a distinct function (compression or generation) as shown in Fig~\ref{fig:datagen} (b).

\paragraph{Compression and Generation.}
Given a document $d_i = \{t_1,\dots,t_m\}$, we append $l$ learnable \textit{memory tokens} $(m_1,\dots,m_l)$ and activate only the \textit{compressor LoRA} $\theta_c$. 
The final-layer hidden states of the memory tokens form the compressed representation:
\[
M_i = \mathrm{LLM}_{\theta_c}([t_1,\dots,t_m,m_1,\dots,m_l])[m{+}1:m{+}l],
\]
which is concatenated with an instruction $I$ to form $T = [I; M_i]$. 
During pretraining, $I$ corresponds to general text-generation tasks (e.g., QA or paraphrasing); during instruction tuning it is replaced with task-specific prompts. 
Only the \textit{generator LoRA} $\theta_g$ is trained via cross-entropy loss:
\begin{equation*}
\mathcal{L}_{\text{CE}}(\theta_c,\theta_g)
= - \sum_{(d_i,I,R_i^*)} \sum_{t=1}^{|R_i^*|}
\log p_{\theta_g}\!\big(a_{i,t}^* \mid I,M_i,a_{i,<t}^*\big).
\label{eq:ce_loss}
\end{equation*}

\paragraph{Compression Alignment.}
To ensure that the compressed representation faithfully reflects the semantics of the original document, we encourage their latent representations to remain aligned.
Intuitively, the memory tokens should summarize the same semantic space as the document tokens, rather than drifting to unrelated regions.
Therefore, we minimize the mean squared error (MSE) between the averaged hidden states of document tokens and memory tokens:\begin{equation}
\mathcal{L}_{\text{MSE}} = 
\left\| 
\frac{1}{|d_i|} \sum_{t \in d_i} h_t 
- 
\frac{1}{l} \sum_{j=1}^{l} h_{m_j}
\right\|_2^2.
\label{mse_loss}
\end{equation}
The total training loss is:
\begin{equation}
\mathcal{L}_{\text{total}} = 
\mathcal{L}_{\text{CE}} + \lambda \mathcal{L}_{\text{MSE}},
\label{eq:total_loss}
\end{equation}
where $\lambda$ balances semantic alignment and generative quality.

\paragraph{Instruction Tuning.}
To adapt the general-purpose pretrained compressor for downstream QA, and also obtain an answer generator that can comprehend the continuous document representation, we optionally performed an additional instruction finetuning training. 
To achieve this, we use downstream training datasets in which the retrieved documents are paired with task instructions.
These document–instruction pairs form the input to the model, while the output is a reference response generated by a teacher model conditioned on the same retrieved documents and instructions.
Similar to the compressor pretraining stage, we jointly finetune the LoRA adapters of both the compressor and the answer generator during this instruction-tuning process.

\begin{figure*}
    \centering
    \includegraphics[width=0.8\linewidth]{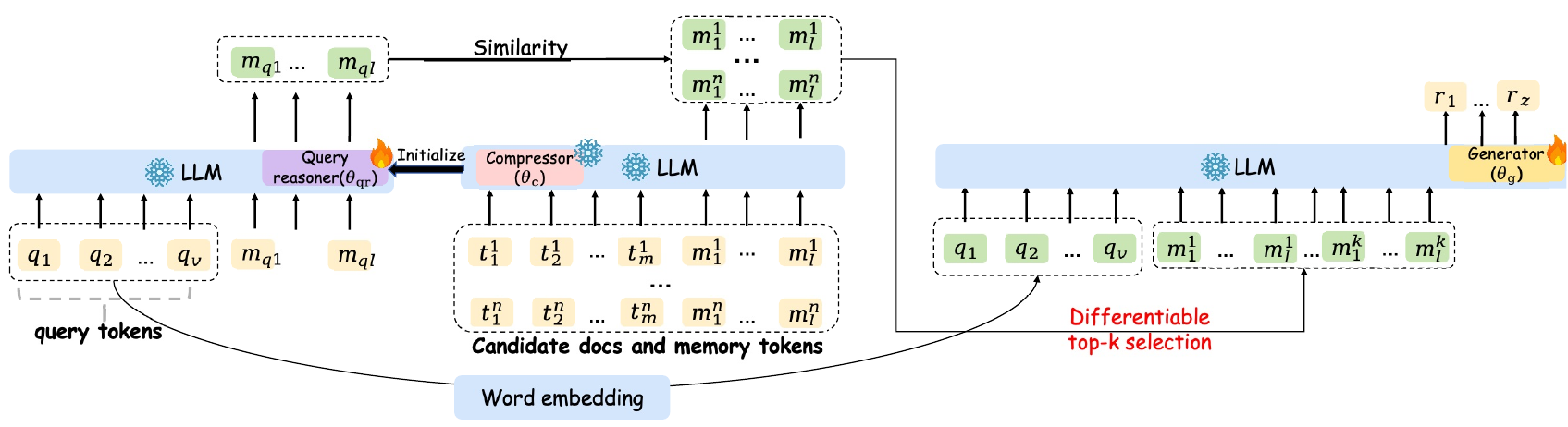}
    \caption{ CLaRa end-to-end training: 
    update query reasoner ($\theta_{qr}$) and generator ($\theta_g$) 
    via language modeling loss using candidate document--question--answer triples.}
    \label{fig:main_method}
    \vspace{-0.3cm}
\end{figure*}

\section{CLaRa: Retrieval and generation joint training}

While the compressor distills documents into compact representations, a key challenge is how to retrieve and leverage these representations effectively for downstream reasoning. Conventional RAG systems train the retriever and generator separately, often resulting in a mismatch between what retrieval considers “relevant’’ and what generation truly needs, as well as two isolated representation spaces. To address this, we propose CLaRa, which unifies retrieval and generation by training both within a single pretrained LLM through a differentiable retrieval module. Achieving end-to-end training, however, requires a retrieval space that is both stable and computationally manageable—full documents are far too large to be re-indexed throughout optimization. To solve this, we use the compressor trained in SCP to produce \textbf{high-quality} compact representations that remain stable even when the rest of the model updates. By retrieving over these frozen compressed vectors, CLaRa can support end-to-end optimization of retrieval and generation using only a shared cross-entropy loss, without requiring relevance-labeled data.

\paragraph{Framework Overview}
As shown in Fig~\ref{fig:main_method}, each document is compressed into a dense embedding $\mathbf{M}_i = \theta_c(t_i)$ using the pretrained compressor $\theta_c$. 
The compressor remains frozen to allow offline document encoding.
We then train a query reasoner ($\theta_{qr}$), a LoRA adapter initialized from $\theta_c$, to represent queries in the same space and with the same number of memory tokens as document representations.
Through NTP training, $\theta_{qr}$ learns not only to encode query intent but also to anticipate relevant document content, enhancing retrieval and answer generation.
For example, given ``\textit{Which city hosted the first modern Olympic Games?}'', an embedding-based retriever may miss ``first'' or ``modern,'' whereas the NTP-trained query reasoner favors documents mentioning ``Athens 1896,''  which better satisfies retrieval with reasoning needs.
We use cosine similarity between query embedding $\mathbf{q}_i$ and document embedding $\mathbf{M}_i$ to obtain a relevance score:
\begin{equation}
s_i = \cos(\mathbf{q}, \mathbf{M}_i), \quad i = 1, \dots, D.
\label{eq_simi}
\end{equation}
The top-$k$ scoring embeddings $\{\mathbf{M}_{1}, \dots, \mathbf{M}_{k}\}$ are concatenated with $\mathbf{q}$ 
and fed to the generator ($\theta_g$), which produces the final answer.
During training, both $\theta_{qr}$ and $\theta_g$ are updated via the unified language modeling loss:
\begin{equation}
\mathcal{L}_{\text{CLaRa}}(\theta_q, \theta_g)
= - \sum_{t=1}^{|R^*|} 
\log p_{\theta_g}\!\left(a_t^* \,\middle|\, \mathbf{q}, 
\mathbf{M}_{(1:k)}, 
a_{<t}^*\right),
\end{equation}
where $R^* = (a_1^*, \dots, a_{|R^*|}^*)$ denotes the reference output. 
Importantly, this allows the retriever (implicitly represented by $\theta_{qr}$) to \textbf{learn through weak supervision from the generation objective}, without explicit reranking labels. Finding real supervised data might be challenging, and our method is data free as it relies only on downstream next token prediction objective to reason on how to retrieve the doc that maximize the likelihood of downstream generation, thus is more flexible and  adaptive.

\paragraph{Differentiable Top-\texorpdfstring{$k$}{k} Selection}
In the CLaRa framework, retrieval and generation are trained jointly, yet their connection is mediated by the top-$k$ selection of relevant documents.  
However, this discrete operation introduces a \textit{broken gradient} problem: the generator’s supervision cannot propagate back to inform the retriever why certain documents should be preferred over others.  

Previous methods such as DDR-RAG~\citep{li2024rag} update the retriever by performing multiple retrieval samplings and using rewards derived from the generator’s outputs under different document samples. However, this RL-based approach suffers from training instability and low efficiency due to repeated sampling. Instead, we use ST estimation for top-$k$ selection, 
which conceptually acts as a ``\textit{soft lens}'' — preserving the discrete retrieval behavior during inference while allowing smooth gradient feedback during training (see Algorithm~\ref{alg:st_topk} in Appendix for details).

Given cosine similarities $s_b = [s_1, \dots, s_D]$ for the $b$th document set in a batch, temperature $\tau$, and masking for previously selected items, the soft and hard selections are defined as:
\begin{equation}
Z_{\text{soft}}[b,j,:] =
\mathrm{softmax}\!\left(
\frac{s_b + \log(\mathrm{mask}_b + \varepsilon)}{\tau}
\right),
\label{eq:soft}
\end{equation}
\begin{equation}
Z_{\text{hard}}[b,j,i] =
\begin{cases}
1, & \text{if } i = \arg\max\nolimits_{i'} Z_{\text{soft}}[b,j,i'], \\
0, & \text{otherwise},
\end{cases}
\label{eq:hard}
\end{equation}
and the final objective that combines the hard and soft representations through a ST estimator is:
\begin{equation}
Z = Z_{\text{hard}} + \big(Z_{\text{soft}} - \mathrm{SG}(Z_{\text{soft}})\big),
\label{eq:st}
\end{equation}
where $\mathrm{SG}(\cdot)$ denotes the stop-gradient operator.  
This maintains discrete behavior in the forward pass while enabling differentiable training through $Z_{\text{soft}}$.  
The aggregated top-$k$ document representation is then computed as:
\begin{equation}
\mathbf{M}^{(k)} = Z M,
\label{eq:topk}
\end{equation}
where $M \in \mathbb{R}^{B\times D\times E}$ is the matrix of all candidate embeddings with B denoting the batch size, D denoting the number of candidate documents,
and E denoting the dimensionality of each document representation, typically defined as
the product of the number of memory tokens and the hidden dimension of the underlying LLM.

\paragraph{Theoretical Justification: Gradient Coupling Analysis}
\label{sec:theory}
We provide an explanation for why learning from NTP yields stronger and more stable training signals for the retriever.
By coupling retrieval and generation through shared representations and ST-based top-$k$ selection, the retriever receives two complementary learning signals.
First, it is encouraged to rank documents that better support correct generation, aligning retrieval probabilities with generation outcomes.
Second, it receives representation-level feedback from the generator, guiding document embeddings toward being more useful for downstream reasoning.
This dual feedback stabilizes joint training by progressively aligning retrieval and generation within a shared semantic space (see Appendix~\ref{app_theo} for details).

\paragraph{Case Study: Query Reasoner~$\theta_{qr}$}

To  probe the information embedded within the query reasoner~$\theta_{qr}$, 
we adopt the \textit{logit lens} analysis technique~\citep{nostalgebraist2020logitlens}. 
For each memory embedding, we project it through the LLM’s output head and record the top-50 tokens with the highest logits as topic tokens. 
We then aggregate and filter these decoded tokens to remove trivial elements such as punctuation or special symbols. 
As shown in Fig.~\ref{fig:logit_lens_examples}, 
for the query \textit{``How many yards did the nephew of Ivory Lee Brown get during his 2004 true freshman season?''}, 
the query embeddings decoded from the reasoner includes the tokens \textit{``NFL'', ``Oklahoma''}, despite the fact that these word do not appear in the question itself. Interestingly, these token \textit{do} occur in the corresponding positive document and serve as a crucial clue for answering the question. This finding indicates that our end-to-end optimization enables the query reasoner to implicitly encode reasoning-relevant knowledge aligned with the gold evidence, 
thus enhancing retrieval accuracy and semantic alignment compared to baseline systems.

\begin{figure}[!htbp]
  \centering
  \scriptsize
\begin{tcolorbox}[
    colback=gray!3,      
    colframe=gray!40,   
    arc=2mm,             
    boxrule=0.4pt,       
    left=3mm, right=3mm, top=2mm, bottom=2mm,
    width=\linewidth,
    title=\textbf{Analysis of Decoded Tokens from Query Reasoner via Logit Lens},
    coltitle=black,
    fonttitle=\bfseries,
]
\scriptsize

\textbf{Question:} How many \textcolor{blue}{yards} did the nephew of \textcolor{blue}{Ivory Lee Brown} get during his \textcolor{blue}{2004} true freshman season? \\[3pt]
\textbf{Reasoned topics from query representation:}  
Truly, Nep, IV, four, yards, \textcolor{red}{NFL}, \textcolor{red}{Oklahoma}, Ned, Neil, Howard, Kentucky... \\[3pt]
\textbf{Retrieved Documents:} 

[1]...\textcolor{blue}{Adrian Lewis Peterson} (born March 21, 1985) is an American football running back for the New Orleans Saints of the \textcolor{red}{National Football League (NFL)}. He played college football at \textcolor{red}{Oklahoma} and was drafted by the Minnesota Vikings seventh overall in the 2007 NFL Draft. Peterson set the \textcolor{blue}{NCAA freshman rushing record with 1,925 yards} as a true freshman during the \textcolor{blue}{2004 season}...

[2]...\textcolor{blue}{Ivory Lee Brown} (born August 17, 1969) is a former professional American football running back in the \textcolor{red}{National Football League} and World League of American Football. He played for the Phoenix Cardinals of the NFL and the San Antonio Riders of the WLAF. Brown is the \textcolor{blue}{uncle of Minnesota Vikings running back Adrian Peterson}... \\[3pt]
...

\textbf{Answer:} \textbf{1,925 yards}
\end{tcolorbox}

\caption{Analysis of Decoded Tokens from Query Reasoner via Logit Lens. 
The highlighted tokens (\textcolor{red}{red}) denote the new information reasoned by the query reasoner, 
while (\textcolor{blue}{blue}) denotes key evidence for solving this multihop task.}
\label{fig:logit_lens_examples}
\end{figure}

\section{Experiments}
\subsection{Experimental setup}

\paragraph{Datasets}
We train the compressor on our curated synthetic data, perform compressor instruction tuning using the training data released by COCOM, and conduct end-to-end training on the corresponding training sets provided with the evaluation tasks. For evaluation, following prior work~\cite{shi2025directretrievalaugmentedoptimizationsynergizing}, we evaluate both the compressor and the end-to-end framework on the full development sets of four widely used question answering benchmarks: \textbf{NQ}~\citep{kwiatkowski-etal-2019-natural}, \textbf{HotpotQA}~\citep{yang-etal-2018-hotpotqa}, \textbf{MuSiQue}~\citep{trivedi-etal-2022-musique}, and \textbf{2WikiMultihopQA}~\citep{ho-etal-2020-constructing}. 

\begin{table*}[!t]
\scriptsize
\centering
\caption{Compressor performance on four QA datasets measured by CEM (\%) (\ref{eva_metric}). DF denotes document format. Compression rates (CR)  is calculated as the ratio between the length of the raw text document and that of the compressed document.  The best performance is highlighted in bold. We show the absolute performance change (±) of our method under different CR relative to its corresponding best baseline performance.  For all methods where the LLM is not explicitly specified, Mistral-7B is used as the default generator. }
\begin{tabular}{ccccccccc}
\bottomrule
Models & CR  & DF & W/ doc & NQ & HotpotQA & Musique & 2Wiki & Average \\
\hline
\rowcolor{gray!15}
\multicolumn{9}{c}{\textbf{Normal}} \\
\hline
Autocompressor & 1x & Vectors & $\checkmark$ & 17.24 & 14.61 & 3.81 & 19.89 & 13.89 \\
Mistral-7B  & 1x & - & $\times$& 35.01 & 27.55 & 5.38 & 38.45 & 26.6 \\
Mistral-7B &  1x & Raw texts& $\checkmark$ &54.58 & 42.94 & 8.94 & 44.24 & 37.67 \\
Phi4-mini & 1x &  - & $\times$ &18.77 & 21.10 & 4.05 & 30.26 & 18.55 \\
Phi4-mini & 1x & Raw texts&  $\checkmark$  &  48.14 & 37.78 & 8.11 & 35.11 & 32.28 \\
\hline 

llmlingua-2 &  4x & Raw texts&  $\checkmark$ & 47.53 & 37.05 & 9.02 & 44.35 & 34.49 \\
SCP-Mistral-7B & 4x& Vectors & $\checkmark$ & \textbf{57.05}\tiny\textcolor{teal}{+9.52} & \textbf{45.09}\tiny\textcolor{teal}{+8.04} & 10.34\tiny\textcolor{teal}{+1.32} & \textbf{46.94}\tiny\textcolor{teal}{+2.59} & \textbf{39.86}\tiny\textcolor{teal}{+5.37} \\
SCP-Phi4-mini & 4x & Vectors & $\checkmark$ & 53.31\tiny\textcolor{teal}{+5.78} & 42.36\tiny\textcolor{teal}{+5.31} & 8.73\tiny\textcolor{red}{-0.29} & 45.22\tiny\textcolor{teal}{+0.87} & 37.40\tiny\textcolor{teal}{+2.91} \\
\hline 

coconum & 16x & Vectors & $\checkmark$ & 24.12 & 21.48 & 3.52 & 24.48 & 18.40 \\
pcc & 16x & Vectors & $\checkmark$ & 31.38 & 22.29 & 3.43 & 19.47 & 19.14 \\
pisco & 16x & Vectors & $\checkmark$ & 54.39 & 41.94 & 10.09 & 44.88 & 37.83 \\
SCP-Mistral-7B& 16x & Vectors & $\checkmark$ & 55.56\tiny\textcolor{teal}{+1.17} & 43.72\tiny\textcolor{teal}{+1.78} & \textbf{10.55}\tiny\textcolor{teal}{+0.46} & 46.00\tiny\textcolor{teal}{+1.12} & 38.96\tiny\textcolor{teal}{+1.13} \\
SCP-Phi4-mini& 16x & Vectors & $\checkmark$ & 51.96\tiny\textcolor{red}{-2.43} & 40.86\tiny\textcolor{red}{-1.08} & 8.61\tiny\textcolor{red}{-1.48} & 44.27\tiny\textcolor{red}{-0.61} & 36.42\tiny\textcolor{red}{-1.42} \\

\hline 
xrag & 128x& Vectors & $\checkmark$  & 32.35 & 25.16 & 3.64 & 28.79 & 22.48 \\
SCP-Mistral-7B& 128x & Vectors & $\checkmark$ & 53.36\tiny\textcolor{teal}{+21.01} & 41.37\tiny\textcolor{teal}{+16.21} & 10.26\tiny\textcolor{teal}{+6.62} & 46.40\tiny\textcolor{teal}{+17.61} & 37.85\tiny\textcolor{teal}{+15.37} \\
SCP-Phi4-mini & 128x& Vectors & $\checkmark$  & 43.09\tiny\textcolor{teal}{+10.74} & 33.92\tiny\textcolor{teal}{+8.76} & 6.87\tiny\textcolor{teal}{+3.23} & 43.70\tiny\textcolor{teal}{+14.91} & 31.90\tiny\textcolor{teal}{+9.42} \\

\toprule
\end{tabular}
\label{result:compression}
\vspace{-0.3cm}
\end{table*}
\paragraph{Baselines}
For compressor evaluation, we benchmark against both classical and recent methods, including 
\textsc{AutoCompressor} \citep{chevalier-etal-2023-adapting}, \textsc{XRAG}\citep{10.5555/3737916.3741392}, \textsc{COCOM} \citep{10.1145/3701551.3703527}, \textsc{PCC} \citep{dai-etal-2025-pretraining}, \textsc{LLMLingual-2} \citep{pan-etal-2024-llmlingua}, and \textsc{PISCO} \citep{louis-etal-2025-pisco}. 
For reranking, we compare with \textsc{BM25}, \textsc{BGE-Reranker} \citep{bge_m3}, \textsc{RankZephyr-7B} \citep{pradeep2023rankzephyreffectiverobustzeroshot}, \textsc{Setwise} \citep{10.1145/3626772.3657813}, and \textsc{Rank-R1} \citep{zhuang2025rankr1enhancingreasoningllmbased}. 
End-to-end QA results are evaluated against representative RAG systems, including 
prompt-based ( \textsc{In-Context RAG}), 
retrieval-optimized (\textsc{ReComp} \citep{xu2024recomp}, \textsc{DPA-RAG} \citep{dong2025understand}), 
fine-tuned LLMs (\textsc{Self-RAG} \citep{asai2024selfrag}, \textsc{Retrobust} \citep{yoran2024making}, \textsc{ChatQA} \citep{10.5555/3737916.3738409}, \textsc{GenGround} \citep{shi-etal-2024-generate}), 
and jointly optimized models (\textsc{DDR-RAG} \citep{li2024rag}, \textsc{DRO} \citep{shi2025directretrievalaugmentedoptimizationsynergizing}).  
Unlike all baselines operating on raw text, our method is the first to jointly optimize reranking and generation directly over \textbf{compressed representations}.  
Full experimental settings are provided in App.~~\ref{app:ex_setup}. Below, we summarize the key findings, while the complete set of additional experiments can be found in the Appendix, including pretraining data analysis (App.~\ref{app:data-quality} \& \ref{app:data_scale}), training process analysis (App.~\ref{app:train_curve}), fidelity and grounding evaluations (App.~\ref{app:more_analysis}), as well as further module analyses  (App.~\ref{app:fied_ground}).

\subsection{Evaluation of Compression Effectiveness}

We evaluate our \textit{document compressor} under two settings: \textbf{Normal} and \textbf{Oracle}.
In the \textbf{Normal} setting, for each test query, we use the \texttt{BGE-large-en-v1.5} model to retrieve the top-5 documents from \texttt{Wikipedia-2021}. All baselines use the same set of retrieved context documents, which are then post-processed differently by each method.
In the \textbf{Oracle} setting, the annotated positive document is included among the top-5 to isolate compression quality from retrieval noise. Table~\ref{result:compression} summarizes results across compression ratios; see Table~\ref{result:compression1} for full results. Our method consistently outperforms all baselines.
Compared to the best soft compression model \textsc{PISCO}, our model achieves average gains of \textbf{1.13\%} (Normal) and \textbf{5.35\%} (Oracle); over the hard compression baseline \textsc{LLMLingual-2}, improvements reach \textbf{5.37\%} and \textbf{17.31\%}, highlighting stronger semantic preservation.

Surprisingly, our model  exceeds the text-based baseline using uncompressed documents,
with average gains of \textbf{2.36\%} on \texttt{Mistral-7B} and \textbf{6.36\%} on \texttt{Phi-4-min}. This implies that well-trained soft compression can retain essential reasoning information while substantially reducing input length. This may be because the compressed representations filter out irrelevant content and focus the generator on the reasoning-relevant context, leading to better generalization than raw text inputs.
While performance declines at extreme compression (beyond 32× in Oracle), the drop remains moderate under Normal conditions due to weaker document relevance.

\begin{table*}[!h]
\centering
\tiny
\caption{End-to-End QA Performance measured by EM (\%) and F1 (\%) (\ref{eva_metric}). * indicates results reported from the DRO paper. CR denotes compression rate. The highest scores are shown in \textbf{bold}, and the second-best ones are \underline{underlined}. Overall, our method achieves comparable performance while reducing the required context length by 16×.}
\begin{tabular}{c|c|c|c|cc|cc|cc|cc|cc}
\hline

\multirow{2}{*}{Models} & \multirow{2}{*}{CR}  & \multirow{2}{*}{\makecell{Retriever \\ tuning}}  &\multirow{2}{*}{\makecell{Generator \\ tuning}}
& \multicolumn{2}{c|}{NQ}
& \multicolumn{2}{c|}{HotpotQA}
& \multicolumn{2}{c|}{Musique}
& \multicolumn{2}{c}{2Wiki} & \multicolumn{2}{c}{Average}
\\
\cline{5-14}
& && & F1 & EM
  & F1 & EM
  & F1 & EM
  & F1 & EM 
  & F1 & EM
   \\
\hline
\rowcolor{gray!17}
\multicolumn{14}{c}{\textbf{Normal data setting}} \\
In-context RAG*  & 1x& $\times$  & $\times$   & 44.69 & 38.07 & 41.27 & 37.14 & 20.11 & 16.78 & 41.02 & 38.51 & 36.77 & 32.62 \\
RECOMP* & 1x  & $\checkmark$ & $\times$ & 42.67 & 37.47 & 42.72 & 38.72 & 24.96 & 17.34 & 38.26 & 32.17 & 37.15 & 31.43 \\
DPA-RAG* &1x & $\checkmark$ & $\times$  & 44.31&37.29&40.53&37.15&20.36&18.45&39.66&39.02 & 36.22 & 32.98 \\
RetRobust* & 1x & $\times$&  $\checkmark$   & 43.82 & 37.03 & 40.54 & 35.59 & 18.16 & 18.11 & 39.11 & 38.65 & 35.41 & 32.34 \\
ChatQA*  & 1x   & $\times$&  $\checkmark$ & 34.54 & 23.64 & 44.60 & 33.40 & 17.05 & 16.64 & 31.90 & 26.80 & 32.02 & 25.12 \\
GenGround*  & 1x  & $\times$&  $\checkmark$  & 42.31 & 40.60 & 44.71 & \underline{41.27} & \underline{24.36} & \underline{20.77} & 42.58 & 39.61 & 38.49 & \underline{35.56} \\
Self-RAG* & 1x  & $\times$&  $\checkmark$  & 31.63 & 29.74 & 27.30 & 16.30 & 21.50 & 9.43 & 27.33 & 23.52 & 26.94 & 19.75 \\
DDR-RAG*  & 1x  & $\checkmark$&  $\checkmark$ & 28.76 & 40.74 & 35.44 & 31.71 & 10.57&13.54 & 38.40 &35.44 & 28.29 & 30.36 \\
DRO-Mistral-7B*  & 1x  & $\checkmark$&  $\checkmark$& \textbf{51.01} & \textbf{42.41} & \textbf{47.87} & \textbf{40.37} & \textbf{25.32} & \textbf{21.36} & \underline{43.65} & \underline{42.12} & \textbf{41.96} & \textbf{36.56} \\
 \cline{2-14}
\multirow{3}{*}{\makecell{CLaRa-Mistral-7B \\ (Pretraining-initialized)}} & 4x   & $\checkmark$&  $\checkmark$ & 40.62 & 31.21 & 39.53 & 29.54 & 14.53 & 6.16 & 42.59 & 38.49 & 34.32 & 26.35 \\
 & 16x  &   $\checkmark$&  $\checkmark$& 41.75 & 32.24 & 44.37 & 33.72 & 15.36 & 6.99 & 43.47 & 39.50 & 36.24 & 28.11 \\
& 32x   &   $\checkmark$&  $\checkmark$&40.68 & 31.36 & 41.84 & 31.26 & 15.32 & 6.66 & 43.23 & 38.98 & 35.27 & 27.06 \\
 \cline{2-14}
\multirow{3}{*}{\makecell{CLaRa-Mistral-7B \\ (Instruction-initialized)}} & 4x & $\checkmark$&  $\checkmark$ &  48.21 & 38.16 & 45.93 & 35.12 & 17.49 &  8.11 & \textbf{47.18} &  \textbf{43.11} & 39.70 & 31.12 \\
& 16x  &  $\checkmark$&  $\checkmark$& \underline{50.89} &  \underline{41.02} & \underline{47.62} & 36.67 & 18.01 & 8.44 & 44.66 &  40.48 & \underline{40.30} & 31.65 \\
 & 32x   &   $\checkmark$&  $\checkmark$&49.72 &  39.88 & 45.73 & 34.85 & 16.83 &7.82 & 42.57 &  38.41 & 38.71 & 30.24 \\

\hline
\rowcolor{gray!17}
\multicolumn{14}{c}{\textbf{Oracle data setting}} \\

 \multirow{3}{*}{\makecell{CLaRa-Mistral-7B \\ (Pretraining-initialized)}} &  4x  & $\checkmark$&  $\checkmark$& 77.80 & 70.52 & 77.66 & 64.83 & 41.59 & 30.33 & 73.20 & 69.14 & 67.56 & 58.70 \\

 & 16x   &   $\checkmark$&  $\checkmark$&73.81 & 65.74 & 69.57 & 56.76 & 31.15 & 21.18 & 65.90 & 61.31 & 60.11 & 51.25 \\

 & 32x   &   $\checkmark$&  $\checkmark$&72.03 & 63.65 & 70.91 & 57.07 & 33.40 & 22.22 & 66.32 & 61.12 & 60.66 & 51.02 \\

  \cline{2-14}

    \multirow{3}{*}{\makecell{CLaRa-Mistral-7B \\ (Instruction-initialized)}} &  4x  & $\checkmark$&  $\checkmark$ & 75.63 &  67.64 & 69.66 &56.92 & 33.19 & 22.42 & 73.86 & 69.74 & 63.08 & 54.18 \\

 & 16x   &  $\checkmark$&  $\checkmark$& 71.54 & 63.29 & 71.17  & 57.54 & 30.77 &  20.56 & 60.37 &  55.73 & 58.46 & 49.28 \\

 & 32x   &  $\checkmark$&  $\checkmark$& 69.75 & 65.17 &  68.87 & 55.20 & 28.87 & 18.45 & 64.38 &  59.32 & 57.97 & 49.53 \\

\toprule
\end{tabular}
\label{end-end-qa-mistral}
\vspace{-0.3cm}
\end{table*}

\subsection{Joint Training Results}

For end-to-end learning, we evaluate our model under both \textbf{Normal} and \textbf{Oracle} settings.
In the Normal setup, each query retrieves the top-20 documents from \texttt{Wikipedia-2021};
the Oracle setup adds annotated positives to the 20-document pool to isolate generation quality from retrieval noise. We compare two initialization strategies for joint reranking–generation training: 
(i) from the compression pretraining checkpoint, and  
(ii) from the instruction-tuned compressor.  
Results are shown in Table~\ref{end-end-qa-mistral}, with full results in Table~\ref{app:end-end-qa-mistral}  in Appendix.

Under the Normal setting, performance remains stable across compression ratios, peaking at \textbf{16–32×}. As 4× might be harder to optim w/ NTP,  
\textsc{CLaRa-Mistral-7B} with 16x surpasses the text-based \textsc{DRO-Mistral-7B}, improving F1 from \textbf{51.01→51.41} on \textbf{NQ} and \textbf{43.65→47.18} on \textbf{2Wiki}.
In the Oracle setting, performance rises notably—F1 exceeds \textbf{75\%} on both \textbf{NQ} and \textbf{HotpotQA}—showing that joint optimization effectively exploits accurate retrieval.

Instruction-tuned initialization outperforms pretraining-based initialization under Normal conditions, especially on NQ and HotpotQA, indicating stronger alignment between compression and answering. However, the gap narrows in the Oracle setting, suggesting initialization matters less when retrieval is reliable.  
Overall, \textsc{CLaRa} demonstrates robust and scalable performance across retrieval qualities and compression ratios.

\begin{figure}
    \centering
    \includegraphics[width=1\linewidth]{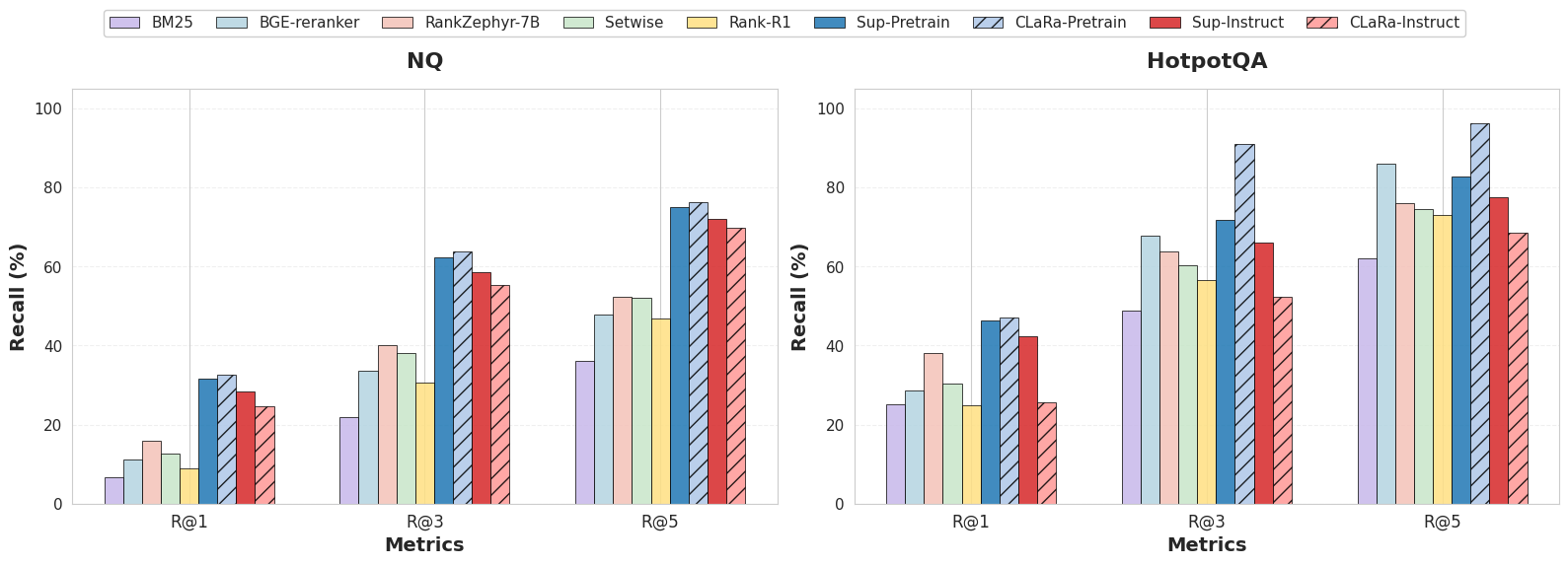}
    \caption{Retrieval performance (Recall@1/3/5) on the \textsc{Mistral-7B} model across different reranking methods under compression ratios = 4 and  various initialization settings on NQ and HotpotQA datasets. 
\textsc{Sup-} denotes models trained with labeled data using contrastive learning for the reranker. \textsc{-Pretrain} denotes experiments conducted using the model checkpoint obtained after pretraining, while \textsc{-Instruct} denotes experiments conducted using the model checkpoint obtained after instruction tuning.}
    \label{retrieval_results_mistral}
    \vspace{-0.4cm}
\end{figure}

\subsection{Retrieval performance}
\label{sec_retrieval}

We evaluate our method on the \textit{document reranking} task to assess retrieval effectiveness under the \textbf{Oracle} setting, where positive documents are guaranteed in the candidate set, allowing accurate computation of \textbf{Recall@k}..
To compare supervision levels, we introduce a fully supervised retriever baseline, \textsc{Sup-Instruct}, which fine-tunes the \textit{Query Reasoner} via contrastive learning with annotated positive and negative documents.
In contrast, our method trains the retriever in a \textbf{weakly supervised} manner, only using the next token prediction loss from the downstream generation. \textbf{Notably, our method does not rely on any supervised data of annotated document relevance labels. }

As shown in Fig.~\ref{retrieval_results_mistral} (full results in Table~\ref{app:retrieval_results_mistral}),
\textsc{CLaRa-Mistral-7B} initialized from pretraining consistently outperforms its instruction-tuned version, indicating that instruction tuning, while improving answer generation, biases the model toward localized evidence at the cost of global semantics crucial for retrieval.

Remarkably, under the pretraining-initialized setup, \textsc{CLaRa} \textbf{even surpasses} the fully supervised \textsc{Sup-Instruct} using ground-truth relevance labels.
On \textbf{HotpotQA} (compression ratio 4), it achieves a \textbf{Recall@5} of \textbf{96.21\%}, exceeding the strongest supervised baseline \textsc{BGE-Reranker} (85.93\%) by \textbf{+10.28\%}.
Despite relying solely on weak generation supervision, \textsc{CLaRa} presumably captures deep semantic correlations between queries and documents and adapts to the downstream scenarios, achieving retrieval quality on par with or surpassing fully supervised models.

\begin{table*}[!t]
\centering
\scriptsize
\caption{Effect of pretraining data composition on instruction-tuning performance
under Oracle (gold context) settings under the 32 compression ratio. We report the absolute score change (±) for each pretraining data setting relative to the No-Pretrain baseline.}
\label{data_composition}
\begin{tabular}{ccccccc}
\hline
Models & Data composition & NQ & HotpotQA& Musique& 2Wikiqa & Average \\
\hline
\multirow{5}{*}{Mistral-7B} & No-pretrain & 70.01 & 61.13 & 29.00 & 57.43 & 54.39 \\
 & SimpleQA & 72.66{\tiny\textcolor{teal}{+2.65}} & 66.41{\tiny\textcolor{teal}{+5.28}} & 35.29{\tiny\textcolor{teal}{+6.29}} & 61.22{\tiny\textcolor{teal}{+3.79}} & 58.90{\tiny\textcolor{teal}{+4.51}} \\
 & Para & 73.86{\tiny\textcolor{teal}{+3.85}} & 68.64{\tiny\textcolor{teal}{+7.51}} & 36.86{\tiny\textcolor{teal}{+7.86}} & 63.22{\tiny\textcolor{teal}{+5.79}} & 60.64{\tiny\textcolor{teal}{+6.25}} \\
 & SimpleQA+ComplexQA & 74.34{\tiny\textcolor{teal}{+4.33}} & 69.31{\tiny\textcolor{teal}{+8.18}} & 36.70{\tiny\textcolor{teal}{+7.70}} & 63.71{\tiny\textcolor{teal}{+6.28}} & 61.02{\tiny\textcolor{teal}{+6.63}} \\
 & SimpleQA+ComplexQA+Para & 73.77{\tiny\textcolor{teal}{+3.76}} & 69.51{\tiny\textcolor{teal}{+8.38}} & 38.31{\tiny\textcolor{teal}{+9.31}} & 64.54{\tiny\textcolor{teal}{+7.11}} & 61.53{\tiny\textcolor{teal}{+7.14}} \\
\hline
\multirow{5}{*}{Phi4-mini} & No-pretrain & 65.54 & 60.32 & 27.31 & 56.39 & 52.39 \\
 & SimpleQA & 68.70{\tiny\textcolor{teal}{+3.16}} & 64.60{\tiny\textcolor{teal}{+4.28}} & 30.41{\tiny\textcolor{teal}{+3.10}} & 57.46{\tiny\textcolor{teal}{+1.07}} & 55.29{\tiny\textcolor{teal}{+2.90}} \\
 & Para & 67.90{\tiny\textcolor{teal}{+2.36}} & 64.72{\tiny\textcolor{teal}{+4.40}} & 31.11{\tiny\textcolor{teal}{+3.80}} & 58.67{\tiny\textcolor{teal}{+2.28}} & 55.60{\tiny\textcolor{teal}{+3.21}} \\
 & SimpleQA+ComplexQA & 69.33{\tiny\textcolor{teal}{+3.79}} & 65.15{\tiny\textcolor{teal}{+4.83}} & 31.15{\tiny\textcolor{teal}{+3.84}} & 57.94{\tiny\textcolor{teal}{+1.55}} & 55.89{\tiny\textcolor{teal}{+3.50}} \\
 & SimpleQA+ComplexQA+Para & 69.90{\tiny\textcolor{teal}{+4.36}} & 65.32{\tiny\textcolor{teal}{+5.00}} & 31.77{\tiny\textcolor{teal}{+4.46}} & 58.52{\tiny\textcolor{teal}{+2.13}} & 56.38{\tiny\textcolor{teal}{+3.99}} \\
\toprule
\end{tabular}
\vspace{-0.3cm}
\end{table*}

\section{Ablation Study}

\paragraph{Pretraining Data Mix}

Each document in our setup is paired with two output types: (i) QA-style question–answer pairs and (ii) paraphrased documents.
To assess the impact of data composition, we vary pretraining objectives and report results in Tables~\ref{data_composition} and \ref{data_composition1}.
For both \textsc{Mistral-7B} and \textsc{Phi4-mini}, using either \textit{SimpleQA} or \textit{Paraphrase} alone already outperforms the no-pretraining baseline, showing that factual reasoning and paraphrastic rewriting both enrich compressed representations.
Combining multiple QA types (\textit{SimpleQA+ComplexQA}) or adding paraphrases (\textit{SimpleQA+ComplexQA+Para}) achieves the best performance, confirming that diverse objectives enhance semantic coverage and generalization—especially under the \textbf{Oracle} setting, where high-quality retrieval amplifies pretraining benefits.

\paragraph{Effect of $\mathcal{L}_{\text{MSE}}$}


\begin{table}[!t]
\centering
\scriptsize
\caption{Instruction-tuning performance with and without MSE loss under different compression ratios and oracle retrieval settings.}
\label{tab:mse_loss}
\begin{tabular}{cclllll}
\bottomrule
Models & CR & NQ & HotpotQA& Musique& 2Wikiqa\\
\hline
Mistral-7B & 32x & 74.65 & 69.05 & 37.32 & 62.98  \\
\qquad w/ mse & 32x & 73.77{\tiny\textcolor{red}{-0.88}} & 69.51{\tiny\textcolor{teal}{+0.46}} & 38.31{\tiny\textcolor{teal}{+0.99}} & 64.54{\tiny\textcolor{teal}{+1.56}} \\
Mistral-7B & 128x & 71.24 & 62.26 & 29.29 & 57.87  \\
\qquad w/ mse & 128x & 69.96{\tiny\textcolor{red}{-1.28}} & 62.09{\tiny\textcolor{red}{-0.17}} & 30.86{\tiny\textcolor{teal}{+1.57}} & 59.08{\tiny\textcolor{teal}{+1.21}} \\
\hline
\end{tabular}
\vspace{-0.4cm}
\end{table}

We analyze the effect of the MSE loss (Eq.~\ref{mse_loss}), which aligns compressed and original document representations.
As Tables~\ref{tab:mse_loss} and \ref{tab:mse_loss1} show, including this loss leads to a modest (0.3–0.6 points on average) but consistent improvement across datasets, confirming that it facilitates semantic preservation during compression. To provide a qualitative perspective, we visualize 4K document embeddings and their corresponding compressed representations using t-SNE (Fig~\ref{fig:mse_loss} in Appendix). 
Without the MSE loss, the two distributions are clearly separated, reflecting a weak correspondence between the memory-token and document spaces. With the MSE loss, the compressed embeddings exhibit strong overlap with the original document representations, demonstrating that the alignment objective effectively enforces semantic consistency between embedding spaces.

\section{Related Work}

\subsection{Embedding-based/Soft Compression}
Recent studies have leveraged LLMs to compress lengthy RAG documents into continuous embeddings for QA tasks~\citep{chevalier-etal-2023-adapting,ge2024incontext,10.5555/3666122.3666970,xiao2025metaembedscalingmultimodalretrieval,dai-etal-2025-pretraining,kuratov-etal-2025-cramming}. Generally, they shorten contexts using continuous representations but are trained independently of LLMs and do not support retrieval–generation co-optimization.
\citet{10.5555/3737916.3741392} propose a projection module mapping each document to a single-vector representation while freezing encoder and decoder parameters, achieving high compression but losing fine-grained semantics essential for RAG. 
\citet{louis-etal-2025-pisco} introduce \textit{PISCO}, which replaces documents with variable \textit{memory-token} representations and jointly trains the encoder and decoder for tighter coupling between compression and generation.
While they suggest pretraining offers limited gains with sufficient instruction data, our results show a more targeted pretraining objective can still yield richer and more informative representations.
The most related work, \citet{louis2025oscaronlinesoftcompression}, jointly trains a query-aware compression model that also functions as a retriever. However, requiring re-compression per query contradicts the goal of reusable, query-independent representations and increases latency. In contrast, our approach enables efficient, fully \textbf{label-free} retriever learning.

\subsection{End-to-End Optimization for RAG}
\textbf{Reinforcement learning approaches} \citep{shi2025directretrievalaugmentedoptimizationsynergizing} allow joint optimization but are unstable and computationally heavy, still relying on raw text.
\textbf{Differentiable reranking} \citep{huang-etal-2025-gumbel} enables gradient-based selection via Gumbel-softmax but likewise processes full documents at every step, leaving the representation mismatch and context length issues unresolved.

As motivated earlier, joint training of retrieval and generation in RAG systems is hindered by the non-differentiability of discrete document selection.
In typical QA pipelines, the retriever reorders retrieved documents before generation~\citep{yu2024rankrag,dong2024dontforgetconnectimproving}, but discrete sampling operations prevent gradient backpropagation.
In contrast, CLaRa uniquely combines compression and joint training: by employing \textbf{length-flexible compressed vectors} in a shared latent space, we enable efficient differentiable selection while drastically reducing context length. Optimized solely through the generator's language modeling loss, CLaRa ensures consistent training–inference alignment and efficient end-to-end learning without explicit retrieval supervision.

\section{Conclusion}
In this paper, we address the challenge of compressing documents into high-quality implicit representations 
to enhance the performance of retrieval-augmented generation (RAG) systems that rely on document embeddings for question answering. 
To this end, we design multiple pretraining objectives that leverage LLM prompting to construct diverse supervision signals, 
including QA pairs—covering both simple and compositional reasoning—and paraphrased documents, 
encouraging the compressor to retain essential semantic information. 
We further introduce an efficient end-to-end training framework that unifies document representations 
across the reranking and generation stages, leading to substantial improvements in retrieval accuracy and answer quality. 
Extensive experiments on multiple QA benchmarks demonstrate that embedding-based contextual compression 
not only reduces input length and computation cost but also bridges the gap between retrieval and generation, 
enabling a more unified and semantically coherent RAG paradigm.

\section*{Impact Statement}


This work aims to advance research in retrieval-augmented generation (RAG) by exploring compression-based representations that unify document understanding and generation. By enabling more compact and semantically dense representations, our approach has the potential to improve the efficiency and scalability of information access systems, particularly in settings where computational or memory resources are constrained.

From a societal perspective, improved retrieval and representation efficiency may benefit applications such as scientific search, education, and knowledge-intensive assistants by reducing latency and resource consumption while maintaining answer quality. At the same time, our current study relies on compressors pretrained primarily on Wikipedia data, which may reflect existing coverage biases in large-scale public corpora. Future work that incorporates more diverse and domain-specific data sources may help mitigate such limitations and improve robustness across modalities and application domains.

We do not identify any immediate negative ethical risks uniquely introduced by our method beyond those commonly associated with retrieval-based language models, such as potential biases inherited from training data or downstream misuse of generated content. As research progresses toward reasoning-oriented or agentic RAG systems that rely on compressed internal representations, it will be important to further examine transparency, controllability, and failure modes in complex decision-making settings.

Overall, this work contributes to ongoing efforts in machine learning to develop more efficient, generalizable, and reasoning-capable systems. We believe its broader societal impacts are aligned with well-established directions in the field and do not raise new ethical concerns that require special consideration at this stage.

\bibliography{custom}
\bibliographystyle{arxiv}

\appendix
\onecolumn

\section{Gradients for Non-shared vs.\ Shared Representations in RAG}
\label{app_theo}
\paragraph{Step 1:}
Let $s_{xd}$ be the retrieval score for query $x$ and document $d$, and let
\begin{align}
\pd = \frac{\exp(\sd)}{\sum_{d'\in C}\exp(s_{xd'})}, \qquad
\px = \sum_{d\in C} \pd\,\qyd, \qquad
\mathcal{L} = -\log \px. \label{eq:step1}
\end{align}

\paragraph{Step 2: product rule inside the sum.}
\begin{align}
\frac{\partial}{\partial s_{xd}}
\Big(\sum_{d'} p(d'|x)\,p(y|x,d')\Big)
&= \sum_{d'} 
\underbrace{\frac{\partial p(d'|x)}{\partial s_{xd}}}_{\text{softmax Jacobian}}
\,p(y|x,d')
+ \sum_{d'} p(d'|x)\,\frac{\partial p(y|x,d')}{\partial s_{xd}}.
\label{eq:step2}
\end{align}

\paragraph{Step 3: softmax Jacobian.}
For $p(d'|x)=\frac{e^{s_{xd'}}}{\sum_j e^{s_{xj}}}$,
\[
\frac{\partial p(d'|x)}{\partial s_{xd}}
= p(d'|x)\big(\mathbf 1[d'=d]-p(d|x)\big).
\]
Therefore
\begin{align}
\sum_{d'} 
\frac{\partial p(d'|x)}{\partial s_{xd}}\;p(y|x,d')
&= \sum_{d'} p(d'|x)\big(\mathbf 1[d'=d]-p(d|x)\big)\,p(y|x,d') \\
&= p(d|x)\,p(y|x,d) - p(d|x)\sum_{d'} p(d'|x)\,p(y|x,d') \\
&= p(d|x)\,\big(p(y|x,d)-p(y|x)\big).
\label{eq:softmax-part}
\end{align}

\paragraph{Step 4: put together.}
Plugging \eqref{eq:softmax-part} into \eqref{eq:step2} and then \eqref{eq:step1} gives
\begin{align}
\frac{\partial \mathcal L}{\partial s_{xd}}
&= -\frac{1}{p(y|x)}\left[
\underbrace{p(d|x)\big(p(y|x,d)-p(y|x)\big)}_{\text{(I) probability path}}
\;+\;
\underbrace{\sum_{d'} p(d'|x)\,\frac{\partial p(y|x,d')}{\partial s_{xd}}}_{\text{(II) representation/generation path}}
\right].
\label{eq:unified-sum}
\end{align}

\paragraph{Step 5: common simplification (assumption).}
If the generator’s conditional $p(y|x,d')$ depends on $s_{xd}$ \emph{only when $d'=d$}
(e.g., each conditional uses its own selected document; non-shared case gives it zero),
then the second sum reduces to a single term:
\[
\sum_{d'} p(d'|x)\,\frac{\partial p(y|x,d')}{\partial s_{xd}}
\;=\; p(d|x)\,\frac{\partial p(y|x,d)}{\partial s_{xd}}.
\]
Under this widely-used assumption, \eqref{eq:unified-sum} becomes
\begin{align}
\frac{\partial \mathcal{L}}{\partial \sd}
= -\frac{1}{\px}\Big[
\underbrace{\pd\big(\qyd - \px\big)}_{\text{(I) probability path}}
+
\underbrace{\pd\,\frac{\partial \qyd}{\partial \sd}}_{\text{(II) representation/generation path}}
\Big].
\label{eq:unified}
\end{align}

\paragraph{Remark (more general shared-conditioning).}
If the generator conditions on a mixture
$r=\sum_j \pi_j(s)\,z_j$ (so every $p(y|x,d')$ shares the \emph{same} $r$),
then $\frac{\partial p(y|x,d')}{\partial s_{xd}}$ is the same for all $d'$ and
$\sum_{d'} p(d'|x)\,\frac{\partial p(y|x,d')}{\partial s_{xd}}=\frac{\partial p(y|x,r)}{\partial s_{xd}}$.
Both forms are consistent; the boxed formula corresponds to the per-document conditional view.

Term (II) is present if the generator's conditional $\qyd$ depends (directly or indirectly) on $\sd$.

\subsection*{Case A: Non-shared representations (retriever $\neq$ generator)}
Here the generator consumes raw tokens or an independent encoder, hence $\qyd$ does not depend on $\sd$:
\begin{align}
\frac{\partial \qyd}{\partial \sd}=0.
\end{align}
Plugging into \eqref{eq:unified} gives the complete gradient:
\begin{align}
\frac{\partial \mathcal{L}}{\partial \sd}
= -\frac{1}{\px}\;\pd\,\big(\qyd - \px\big).
\label{eq:nonshared}
\end{align}
This expression already accounts for the softmax coupling via $\px$ and is more accurate than writing only $-\frac{1}{\px}\,\qyd\,\partial \pd/\partial \sd$.

\subsection*{Case B: Shared representations (retriever $=$ generator)}
When retriever and generator share embeddings, $\qyd$ depends on $\sd$ through the generator's conditioning vector. A common differentiable conditioning is
\begin{align}
\pi_j \;=\; \frac{\exp(s_{xj}/\tau)}{\sum_{\ell}\exp(s_{x\ell}/\tau)},
\qquad
r \;=\; \sum_{j\in C} \pi_j\, z_j,
\qquad
\qyd \equiv p(y\,|\,x,r),
\label{eq:mixture}
\end{align}
where $\tau>0$ is the temperature, $z_j$ are document embeddings, and $r$ is fed to the generator.

Let
\begin{align}
g \;\triangleq\; \frac{\partial \log p(y\,|\,x,r)}{\partial r}
\;=\; \sum_{t}\frac{\partial \log p(y_t\,|\,y_{<t},x,r)}{\partial r}.
\label{eq:gdef}
\end{align}
Using the softmax Jacobian,
\begin{align}
\frac{\partial r}{\partial \sd}
= \sum_{j}\frac{\partial \pi_j}{\partial \sd} z_j
= \frac{1}{\tau}\,\pi_d\,(z_d - r).
\label{eq:drds}
\end{align}
By chain rule,
\begin{align}
\frac{\partial \qyd}{\partial \sd}
= p(y\,|\,x,r)\; g^\top \frac{\partial r}{\partial \sd}
= \frac{p(y\,|\,x,r)}{\tau}\;\pi_d\; g^\top\big(z_d - r\big).
\label{eq:second-term}
\end{align}
Substituting \eqref{eq:second-term} into \eqref{eq:unified} yields the full shared-representation gradient:
\begin{align}
\frac{\partial \mathcal{L}}{\partial \sd}
= -\frac{1}{\px}\Big[
\pd\,\big(\qyd - \px\big)
\;+\;
\frac{p(r\,|\,x)\,p(y\,|\,x,r)}{\tau}\;\pi_d\; g^\top\big(z_d - r\big)
\Big].
\label{eq:shared-full}
\end{align}

\paragraph{Straight-through (ST) note.}
If the forward pass uses hard top-$k$ selection (argmax/indices) but the backward pass adopts the softmax gradient (ST estimator), then formulas
\eqref{eq:drds}--\eqref{eq:shared-full} remain the correct backpropagation rules (with $\pi$ computed from the scores for the backward pass).

\subsection*{Optional: Cosine-similarity score backpropagation}
If the score is cosine similarity
\begin{align}
\sd \;=\; \frac{q^\top z_d}{\|q\|\,\|z_d\|},
\end{align}
then the required Jacobians are
\begin{align}
\frac{\partial \sd}{\partial q}
&=
\frac{1}{\|q\|\,\|z_d\|}
\left(
z_d
-
\sd\,\frac{q}{\|q\|^2}\,\|z_d\|
\right),
&
\frac{\partial \sd}{\partial z_d}
&=
\frac{1}{\|q\|\,\|z_d\|}
\left(
q
-
\sd\,\frac{z_d}{\|z_d\|^2}\,\|q\|
\right).
\end{align}
Hence
\begin{align}
\frac{\partial \mathcal{L}}{\partial q}
=\sum_{d\in C}\frac{\partial \mathcal{L}}{\partial \sd}\,\frac{\partial \sd}{\partial q},
\qquad
\frac{\partial \mathcal{L}}{\partial z_d}
=\frac{\partial \mathcal{L}}{\partial \sd}\,\frac{\partial \sd}{\partial z_d}.
\end{align}

\begin{algorithm}[t]
\footnotesize
\caption{Differentiable Top-$k$ Selection with Straight-Through Estimator in CLaRA}
\label{alg:st_topk}
\begin{algorithmic}[1]
\State \textbf{Input:} Similarity scores $s \in \mathbb{R}^{B \times D}$, temperature $\tau$, number of selections $k$
\State \textbf{Output:} Selection tensor $Z \in \mathbb{R}^{B \times k \times D}$ and top-$k$ indices $\{\text{r}_j\}_{j=1}^k$
\vspace{2pt}
\State $\tilde{s} \gets s / \max(\tau, 10^{-6})$
\State Initialize $Z_{\text{hard}}, Z_{\text{soft}} \gets \mathbf{0}^{B\times k\times D}$, \; $\text{taken} \gets \mathbf{0}^{B\times D}$
\For{$j = 1$ \textbf{to} $k$}
    \State \textbf{(1) Hard selection:} 
        $r_j \gets \arg\max_i \tilde{s}(:, i)$ on unmasked candidates
    \State \hspace{2.5em} $Z_{\text{hard}}[:, j, r_j] \gets 1$
    \State \textbf{(2) Soft selection:}
        $\text{mask} \gets 1 - \mathrm{SG}(\text{taken})$
    \State \hspace{2.5em} $\text{logits}_j \gets \tilde{s} + \log(\text{mask} + \varepsilon)$
    \State \hspace{2.5em} $p_j \gets \mathrm{softmax}(\text{logits}_j)$
    \State \hspace{2.5em} $Z_{\text{soft}}[:, j, :] \gets p_j$
    \State \hspace{2.5em} $\text{taken} \gets \min(\text{taken} + Z_{\text{hard}}[:, j, :], 1)$
\EndFor
\State \textbf{(3) Straight-through estimator:} 
    $Z \gets Z_{\text{hard}} + (Z_{\text{soft}} - \mathrm{SG}(Z_{\text{soft}}))$
\State \textbf{Return} $(Z, \{ r_j \}_{j=1}^k)$
\end{algorithmic}
\end{algorithm}

\begin{table}[!t]
\footnotesize
\centering
\caption{Compressor performance on four QA datasets. The best performance is highlighted in bold. We show the absolute performance change (±) of our method under different compression rates relative to its corresponding w/ retrieval setting. CR denotes compression rate.
 }
\label{result:compression1}
\begin{tabular}{ccccccc}
\bottomrule
Models & CR & NQ & HotpotQA & Musique & 2Wiki & Average \\
\hline
\rowcolor{gray!15}
\multicolumn{7}{c}{\textbf{Normal}} \\
\hline
Autocompressor & 1x & 17.24 & 14.61 & 3.81 & 19.89 & 13.89 \\
xrag & 128x & 32.35 & 25.16 & 3.64 & 28.79 & 22.48 \\
coconum & 16x & 24.12 & 21.48 & 3.52 & 24.48 & 18.40 \\
pcc & 16x & 31.38 & 22.29 & 3.43 & 19.47 & 19.14 \\
llmlingua-2 & 4x & 47.53 & 37.05 & 9.02 & 44.35 & 34.49 \\
pisco & 16x & 54.39 & 41.94 & 10.09 & 44.88 & 37.83 \\
\hline
Mistral-7B w/o BGE retrieval & 1x & 35.01 & 27.55 & 5.38 & 38.45 & 26.6 \\
Mistral-7B w/ BGE retrieval & 1x & 54.58 & 42.94 & 8.94 & 44.24 & 37.67 \\
\multirow{6}{*}{SCP-Mistral-7B }& 4x & \textbf{57.05}\tiny\textcolor{teal}{+2.47} & \textbf{45.09}\tiny\textcolor{teal}{+2.15} & \textbf{10.34}\tiny\textcolor{teal}{+1.40} & \textbf{46.94}\tiny\textcolor{teal}{+2.70} & \textbf{39.86}\tiny\textcolor{teal}{+2.19} \\
 & 16x & 55.56\tiny\textcolor{teal}{+0.98} & 43.72\tiny\textcolor{teal}{+0.78} & 10.55\tiny\textcolor{teal}{+1.61} & 46.00\tiny\textcolor{teal}{+1.76} & 38.96\tiny\textcolor{teal}{+1.29} \\
 & 32x & 54.64\tiny\textcolor{teal}{+0.06} & 43.52\tiny\textcolor{teal}{+0.58} & 10.55\tiny\textcolor{teal}{+1.61} & 46.58\tiny\textcolor{teal}{+2.34} & 38.82\tiny\textcolor{teal}{+1.15} \\
 & 64x & 54.18\tiny\textcolor{red}{-0.40} & 42.17\tiny\textcolor{red}{-0.77} & 10.17\tiny\textcolor{teal}{+1.23} & 47.03\tiny\textcolor{teal}{+2.79} & 38.39\tiny\textcolor{teal}{+0.72} \\
 & 128x & 53.36\tiny\textcolor{red}{-1.22} & 41.37\tiny\textcolor{red}{-1.57} & 10.26\tiny\textcolor{teal}{+1.32} & 46.40\tiny\textcolor{teal}{+2.16} & 37.85\tiny\textcolor{teal}{+0.18} \\
 & 256x & 52.84\tiny\textcolor{red}{-1.74} & 40.00\tiny\textcolor{red}{-2.94} & 10.38\tiny\textcolor{teal}{+1.44} & 46.31\tiny\textcolor{teal}{+2.07} & 37.38\tiny\textcolor{red}{-0.29} \\
\cline{2-7}
Phi4-mini w/o BGE retrieval & 1x & 18.77 & 21.10 & 4.05 & 30.26 & 18.55 \\
Phi4-mini w/ BGE retrieval & 1x & 48.14 & 37.78 & 8.11 & 35.11 & 32.28 \\
\multirow{6}{*}{SCP-Phi4-mini } & 4x & \textbf{53.31}\tiny\textcolor{teal}{+5.17} & \textbf{42.36}\tiny\textcolor{teal}{+4.58} & \textbf{8.73}\tiny\textcolor{teal}{+0.62} & \textbf{45.22}\tiny\textcolor{teal}{+10.11} & \textbf{37.40}\tiny\textcolor{teal}{+5.12} \\
 & 16x & 51.96\tiny\textcolor{teal}{+3.82} & 40.86\tiny\textcolor{teal}{+3.08} & 8.61\tiny\textcolor{teal}{+0.50} & 44.27\tiny\textcolor{teal}{+9.16} & 36.42\tiny\textcolor{teal}{+4.14} \\
 & 32x & 49.30\tiny\textcolor{teal}{+1.16} & 38.62\tiny\textcolor{teal}{+0.84} & 7.70\tiny\textcolor{red}{-0.41} & 43.71\tiny\textcolor{teal}{+8.60} & 34.83\tiny\textcolor{teal}{+2.55} \\
 & 64x & 45.72\tiny\textcolor{red}{-2.42} & 35.75\tiny\textcolor{red}{-2.03} & 6.50\tiny\textcolor{red}{-1.61} & 43.96\tiny\textcolor{teal}{+8.85} & 32.98\tiny\textcolor{teal}{+0.70} \\
 & 128x & 43.09\tiny\textcolor{red}{-5.05} & 33.92\tiny\textcolor{red}{-3.86} & 6.87\tiny\textcolor{red}{-1.24} & 43.70\tiny\textcolor{teal}{+8.59} & 31.90\tiny\textcolor{red}{-0.38} \\
 & 256x & 42.73\tiny\textcolor{red}{-5.41} & 34.02\tiny\textcolor{red}{-3.76} & 6.87\tiny\textcolor{red}{-1.24} & 43.75\tiny\textcolor{teal}{+8.64} & 31.84\tiny\textcolor{red}{-0.44} \\
\hline
\rowcolor{gray!15}
\multicolumn{7}{c}{\textbf{Oracle}} \\
\hline
Autocompressor & 1x & 29.47 & 19.24 & 7.16 & 26.74 & 20.65 \\
xrag & 128x & 42.60 & 30.21 & 7.03 & 30.94 & 27.70 \\
coconum & 16x & 25.61 & 21.72 & 3.64 & 24.63 & 18.90 \\
pcc & 16x & 49.62 & 34.56 & 18.25 & 27.56 & 32.50 \\
llmlingua-2 & 4x & 63.99 & 52.42 & 27.47 & 53.92 & 49.45 \\
pisco & 16x & 73.44 & 66.53 & 33.80 & 60.45 & 58.55 \\
\hline
Mistral-7B w/ BGE retrieval & 1x & 71.64 & 70.77 & 45.72 & 68.83 & 64.24 \\
\multirow{6}{*}{SCP-Mistral-7B } & 4x & \textbf{76.50}\tiny\textcolor{teal}{+4.86} & \textbf{73.81}\tiny\textcolor{teal}{+3.04} & \textbf{46.26}\tiny\textcolor{teal}{+0.54} & \textbf{70.48}\tiny\textcolor{teal}{+1.65} & \textbf{66.76}\tiny\textcolor{teal}{+2.52} \\
 & 16x & 75.48\tiny\textcolor{teal}{+3.84} & 70.79\tiny\textcolor{teal}{+0.02} & 43.15\tiny\textcolor{red}{-2.57} & 66.16\tiny\textcolor{red}{-2.67} & 63.90\tiny\textcolor{red}{-0.34} \\
 & 32x & 73.77\tiny\textcolor{teal}{+2.13} & 69.51\tiny\textcolor{red}{-1.26} & 38.31\tiny\textcolor{red}{-7.41} & 64.54\tiny\textcolor{red}{-4.29} & 61.53\tiny\textcolor{red}{-2.71} \\
 & 64x & 71.90\tiny\textcolor{teal}{+0.26} & 66.22\tiny\textcolor{red}{-4.55} & 34.96\tiny\textcolor{red}{-10.76} & 61.55\tiny\textcolor{red}{-7.28} & 58.66\tiny\textcolor{red}{-5.58} \\
 & 128x & 69.96\tiny\textcolor{red}{-1.68} & 62.09\tiny\textcolor{red}{-8.68} & 30.86\tiny\textcolor{red}{-14.86} & 59.08\tiny\textcolor{red}{-9.75} & 55.50\tiny\textcolor{red}{-8.74} \\
 & 256x & 68.82\tiny\textcolor{red}{-2.82} & 59.93\tiny\textcolor{red}{-10.84} & 26.19\tiny\textcolor{red}{-19.53} & 56.50\tiny\textcolor{red}{-12.33} & 52.86\tiny\textcolor{red}{-11.38} \\
\cline{2-7}
Phi4-mini w/ BGE retrieval & 1x & 66.10 & 64.06 & 37.07 & 52.69 & 54.98 \\
\multirow{6}{*}{SCP-Phi4-mini }  & 4x & \textbf{73.67}\tiny\textcolor{teal}{+7.57} & \textbf{72.41}\tiny\textcolor{teal}{+8.35} & \textbf{40.13}\tiny\textcolor{teal}{+3.06} & \textbf{64.22}\tiny\textcolor{teal}{+11.53} & \textbf{62.61}\tiny\textcolor{teal}{+7.63} \\
 & 16x & 73.17\tiny\textcolor{teal}{+7.07} & 70.26\tiny\textcolor{teal}{+6.20} & 38.39\tiny\textcolor{teal}{+1.32} & 63.15\tiny\textcolor{teal}{+10.46} & 61.24\tiny\textcolor{teal}{+6.26} \\
 & 32x & 69.90\tiny\textcolor{teal}{+3.80} & 65.32\tiny\textcolor{teal}{+1.26} & 31.77\tiny\textcolor{red}{-5.30} & 58.52\tiny\textcolor{teal}{+5.83} & 56.38\tiny\textcolor{teal}{+1.40} \\
 & 64x & 64.72\tiny\textcolor{red}{-1.38} & 57.79\tiny\textcolor{red}{-6.27} & 23.54\tiny\textcolor{red}{-13.53} & 53.11\tiny\textcolor{teal}{+0.42} & 49.79\tiny\textcolor{red}{-5.19} \\
 & 128x & 60.44\tiny\textcolor{red}{-5.66} & 51.52\tiny\textcolor{red}{-12.54} & 19.28\tiny\textcolor{red}{-17.79} & 50.29\tiny\textcolor{red}{-2.40} & 45.38\tiny\textcolor{red}{-9.60} \\
 & 256x & 60.12\tiny\textcolor{red}{-5.98} & 51.54\tiny\textcolor{red}{-12.52} & 19.61\tiny\textcolor{red}{-17.46} & 50.33\tiny\textcolor{red}{-2.36} & 45.40\tiny\textcolor{red}{-9.58} \\
\toprule
\end{tabular}
\end{table}

\begin{table}[!htbp]
\centering
\scriptsize
\caption{End-to-End QA Performance. * indicates that the results are reported from the DRO paper. CR means compression rate.}
\label{app:end-end-qa-mistral}
\begin{tabular}{c|c|c|cc|cc|cc|cc}
\hline

\multirow{2}{*}{Models} & \multirow{2}{*}{CR} & \multirow{2}{*}{Retrieval Mode}
& \multicolumn{2}{c|}{NQ}
& \multicolumn{2}{c|}{HotpotQA}
& \multicolumn{2}{c|}{Musique}
& \multicolumn{2}{c}{2Wiki}
\\
\cline{4-11}
& & & F1 & EM
  & F1 & EM
  & F1 & EM
  & F1 & EM
   \\
\hline
\rowcolor{gray!15}
\multicolumn{11}{c}{\textbf{Prompting-based Method}} \\
GenGround*  & 1x & Normal & 42.31 & 40.60 & 44.71 & 41.27 & 24.36 & 20.77 & 42.58 & 39.61   \\
In-context RAG*  & 1x & Normal & 44.69 & 38.07 & 41.27 & 37.14 & 20.11 & 16.78 & 41.02 & 38.51 \\
\hline
\rowcolor{gray!15}
\multicolumn{11}{c}{\textbf{Retrieval tuning}} \\
RECOMP* & 1x & Normal & 42.67 & 37.47 & 42.72 & 38.72 & 24.96 & 17.34 & 38.26 & 32.17 \\
DPA-RAG* &1x &Normal & 44.31&37.29&40.53&37.15&20.36&18.45&39.66&39.02 \\
\hline
\rowcolor{gray!15}
\multicolumn{11}{c}{\textbf{LLM Fine-tuning}} \\
RetRobust* & 1x& Normal & 43.82 & 37.03 & 40.54 & 35.59 & 18.16 & 18.11 & 39.11 & 38.65 \\
ChatQA*  & 1x & Normal & 34.54 & 23.64 & 44.60 & 33.40 & 17.05 & 16.64 & 31.90 & 26.80   \\
Self-RAG* & 1x & Normal & 31.63 & 29.74 & 27.30 & 16.30 & 21.50 & 9.43 & 27.33 & 23.52   \\
\hline
\rowcolor{gray!15}
\multicolumn{11}{c}{\textbf{End-to-end optimization}} \\
DDR-RAG*  & 1x & Normal & 28.76 & 40.74 & 35.44 & 31.71 & 10.57&13.54 & 38.40 &35.44   \\
DRO-Mistral-7B*  & 1x & Normal & 51.01 & 42.41 & 47.87 & 40.37 & 25.32 & 21.36 & 43.65 & 42.12   \\
\hline
\rowcolor{gray!15}
\multicolumn{11}{c}{\textbf{Pretraining-initialized}} \\

\multirow{12}{*}{CLaRa-Mistral-7B} & 4x & \multirow{6}{*}{Normal}  & 40.62 & 31.21 & 39.53 & 29.54 & 14.53 & 6.16 & 42.59 & 38.49\\
 & 16x & & 41.75 & 32.24 & 44.37 & 33.72 & 15.36 & 6.99 & 43.47 & 39.50 \\
& 32x &  & 40.68 & 31.36 & 41.84 & 31.26 & 15.32 & 6.66 & 43.23 & 38.98  \\
& 64x &  & 41.58 & 31.38 & 41.62 & 31.12 & 14.78 & 6.16 & 42.64 & 38.40 \\
& 128x & & 42.04 & 31.78 & 42.26 & 31.78 & 15.53 & 6.37 & 41.80 & 37.37 \\
 & 256x &  & 42.90 & 32.58 & 41.32 & 30.44 & 15.44 & 6.41 & 41.96 & 37.60  \\
 \cline{2-11}
 &  4x& \multirow{6}{*}{Oracle} & 77.80 & 70.52 & 77.66 & 64.83 & 41.59 & 30.33 & 73.20 & 69.14  \\

 & 16x &  & 73.81 & 65.74 & 69.57 & 56.76 & 31.15 & 21.18 & 65.90 & 61.31\\

 & 32x &  & 72.03 & 63.65 & 70.91 & 57.07 & 33.40 & 22.22 & 66.32 & 61.12 \\

 & 64x &  & 68.18 & 59.56 & 67.64 & 53.22 & 28.43 & 17.42 & 62.53 & 57.02\\

 & 128x &  & 68.66 & 59.25 & 66.51 & 52.30 & 28.44 & 16.67 & 64.82 & 58.97 \\

 & 256x  &  & 66.85 & 57.17 & 63.43 & 49.08 & 27.44 & 16.92 & 62.96 & 57.35  \\
\cline{2-11}
 \multirow{12}{*}{CLaRa-Phi4-mini} & 4x & \multirow{6}{*}{Normal} & 39.69 & 30.41 & 37.10 & 27.33 & 15.20 & 6.08 & 38.43 & 34.26  \\
 & 16x &  & 31.93 & 23.38 & 37.21 & 27.22 & 14.30 & 4.84 & 40.03 & 35.62  \\
 & 32x &  & 30.70 & 21.78 & 37.14 & 26.99 & 13.26 & 4.39 & 38.15 & 33.82 \\
 & 64x &  & 28.88 & 19.91 & 34.98 & 25.07 & 13.31 & 4.55 & 37.74 & 33.57 \\
& 128x &  & 29.26 & 19.85 & 34.73 & 24.95 & 13.07 & 4.01 & 36.41 & 32.23 \\
 & 256x &  & 29.92 & 20.53 & 34.10 & 24.62 & 13.07 & 4.22 & 35.98 & 31.61 \\
\cline{2-11}

 & 4x & \multirow{6}{*}{Oracle}  & 61.07 & 52.15 & 59.82 & 46.99 & 25.87 & 15.76 & 56.84 & 52.12 \\

 &16x  &  & 65.09 & 55.96 & 57.87 & 45.26 & 21.09 & 11.25 & 55.75 & 50.41  \\

 & 32x &  & 62.35 & 52.07 & 51.06 & 38.33 & 20.98 & 10.92 & 50.68 & 45.41  \\

 & 64x &  & 58.51 & 47.08 & 55.47 & 41.62 & 23.89 & 12.99 & 53.90 & 48.26 \\

 &  128x&  & 56.13 & 44.52 & 52.68 & 38.49 & 20.74 & 9.85 & 49.97 & 44.11  \\

 & 256x &  & 54.58 & 43.24 & 51.62 & 37.77 & 21.45 & 9.89 & 48.46 & 42.63\\

\hline
\rowcolor{gray!15}
\multicolumn{11}{c}{\textbf{Instruction-tuned-initialized}} \\
\multirow{12}{*}{CLaRa-Mistral-7B} & 4x & \multirow{6}{*}{Normal} & 48.21 & 38.16 & 45.93 & 35.12 & 17.49 &  8.11 & 47.18 &  43.11 \\
& 16x & & 50.89 &  41.02 & 47.62 & 36.67 & 18.01 & 8.44 & 44.66 &  40.48  \\
 & 32x &  & 49.72 &  39.88 & 45.73 & 34.85 & 16.83 &7.82 & 42.57 &  38.41  \\
& 64x &  & 50.91 &  41.07 & 45.68 & 34.74 & 16.76 & 7.65 & 40.34 &  35.91 \\
 & 128x &  & 51.41 &  41.27 & 44.63 &  33.88 & 15.75 & 7.03 & 40.55 &  36.12  \\
 & 256x &  & 50.57 &  40.39 & 43.02 & 32.26 & 16.02 & 6.99 & 40.10 &35.77  \\
  \cline{2-11}

    &  4x& \multirow{6}{*}{Oracle}  & 75.63 &  67.64 & 69.66 &56.92 & 33.19 & 22.42 & 73.86 & 69.74\\

 & 16x &  & 71.54 & 63.29 & 71.17  & 57.54 & 30.77 &  20.56 & 60.37 &  55.73 \\

 & 32x &  & 69.75 & 65.17 &  68.87 & 55.20 & 28.87 & 18.45 & 64.38 &  59.32 \\

 & 64x &  & 68.17 &  59.04 & 66.64 & 52.87 & 27.30 &16.96 & 60.98 &  55.59 \\

 &  128x&  & 66.95 &  57.61 & 64.09 & 54.63 & 26.11 & 15.97 & 62.34 &  56.64 \\

 & 256x &  & 65.60 & 55.65 & 61.79 & 47.74 & 27.67 & 17.05 & 59.40 & 53.75  \\ 
\cline{2-11}
 \multirow{12}{*}{CLaRa-Phi4-mini} & 4x &  \multirow{6}{*}{Normal} & 41.86 &  31.96 & 39.44 & 29.32 & 15.70 & 5.59 & 37.63 & 33.40  \\
 & 16x &  & 42.17 & 32.61 & 42.77 & 32.00 & 15.84 & 6.08 & 36.69 & 32.47 \\
 & 32x &  & 39.14 &  29.45 & 42.59 &  31.47 & 15.55 & 5.71 & 41.47 & 36.68 \\
 & 64x &  & 36.91 &27.09 & 38.90 &  28.02 & 14.08 &  4.88 & 39.52 & 34.98  \\
 & 128x &  & 36.26 &  26.34 & 36.39  & 26.44 & 14.70 &  5.42 & 37.14 &  32.85 \\
 & 256x &  & 37.58 & 27.48 & 35.84 & 25.73 & 13.66 &  4.92 & 36.26 & 32.11 \\
\cline{2-11}
 & 4x & \multirow{6}{*}{Oracle}  & 55.53 & 45.94 & 55.28 &  43.24 & 25.96 & 15.14 & 55.57 &  50.18  \\

 &  16x&  & 58.62 &  48.90 & 56.47 & 43.45 & 23.07 &12.49 & 56.85 & 51.57 \\

 & 32x &  & 61.15 & 50.45 & 56.31  & 43.13 & 21.28 &  11.29 & 51.21 &  45.70 \\

 & 64x &  & 57.63 &  46.62 & 52.39 & 38.94 & 22.38  & 11.63 & 48.11 & 42.83  \\

 & 128x &  & 56.26 & 44.77 & 50.74  & 37.68 & 22.27 & 11.79 & 47.64 &  42.31 \\

 & 256x &  & 54.55 &43.09 & 50.00 &  36.78 & 20.92 &  11.01 & 46.85 & 41.66  \\

\toprule
\end{tabular}
\end{table}

\begin{table}[!htbp]
\centering
\scriptsize
\caption{
Retrieval performance (Recall@1/3/5) on the \textsc{Mistral-7B}  and \textsc{Phi-4-mini} model across different reranking methods under various compression ratios (CR) and initialization settings on four QA datasets. 
\textsc{Sup-} denotes models trained with labeled data using contrastive learning for the reranker.
}
\label{app:retrieval_results_mistral}
\begin{tabular}{c|c|ccc|ccc|ccc|ccc}
\hline
\multirow{2}{*}{Models} & \multirow{2}{*}{CR} 
& \multicolumn{3}{c|}{NQ} 
& \multicolumn{3}{c|}{HotpotQA} 
& \multicolumn{3}{c|}{Musique} 
& \multicolumn{3}{c}{2Wiki} 
 \\
\cline{3-14}
& & R@1 & R@3 & R@5 
  & R@1 & R@3 & R@5 
  & R@1 & R@3 & R@5 
  & R@1 & R@3 & R@5 
\\
\hline
BM25 &1x& 6.57 & 21.89 & 35.99 & 25.12 & 48.86 & 62.09 & 13.00 & 29.12 & 39.13 & 15.23 & 37.06 & 51.40 \\
BGE-reranker &1x &11.18 & 33.56 & 47.78 & 28.53 & 67.91 & 85.93 & 18.32 & 42.45 & 54.13 & 22.21 & 54.95 & 68.32 \\
RankZephyr-7B&1x & 16.00 & 40.20 & 52.40 & 38.00 & 63.70 & 75.90 & 21.04 & 40.61 & 50.65 & 33.40 & 61.55 & 74.69 \\

Setwise & 1x& 12.60 & 38.20 & 52.10 & 30.40 & 60.30 & 74.60 & 18.50 & 40.30 & 51.60 & 26.20 & 57.99 & 71.61 \\

Rank-R1&1x & 8.95 & 30.60 & 46.72 & 24.87 & 56.56 & 73.06 & 16.77 & 38.49 & 51.68 & 24.27 & 59.02 & 77.02 \\
\hline
\rowcolor{gray!15}
\multicolumn{14}{c}{\textbf{Pretraining-initialized}} \\
\multirow{6}{*}{Sup-Mistral-7B} & 4x & 31.57 & 62.34 & 74.96 & 46.33 & 71.76 & 82.84 & 32.35 & 50.30 & 60.02 & 42.38 & 74.51 & 85.02  \\
 & 16x & 30.66 & 61.05 & 73.30 & 44.00 & 69.66 & 80.45 & 29.44 & 45.07 & 54.63 & 40.03 & 65.48 & 77.27  \\
 & 32x & 30.19 & 60.89 & 73.93 & 45.69 & 73.76 & 83.80 & 31.34 & 48.20 & 57.86 & 42.08 & 68.09 & 79.41 \\
 & 64x & 29.88 & 60.38 & 73.53 & 45.92 & 74.35 & 84.46 & 31.32 & 46.91 & 55.93 & 42.13 & 68.26 & 79.52  \\
 & 128x & 29.02 & 59.25 & 72.33 & 44.79 & 73.76 & 83.76 & 29.42 & 45.97 & 55.21 & 40.65 & 65.61 & 76.97 \\
 & 256x & 27.54 & 56.67 & 70.00 & 44.50 & 73.89 & 83.81 & 29.13 & 46.33 & 56.04 & 39.99 & 64.87 & 76.22   \\
\cline{2-14}
\multirow{6}{*}{CLaRa-Mistral-7B} & 4x & 32.62 & 63.71 & 76.38 & 47.07 & 90.90 & 96.21 & 30.45 & 59.99 & 72.46 & 34.37 & 68.08 & 79.13 \\
& 16x & 28.45 & 58.97 & 71.88 & 42.01 & 77.80 & 87.85 & 22.57 & 44.34 & 58.83 & 32.04 & 67.12 & 82.50 \\
 & 32x & 28.06 & 59.68 & 73.21 & 43.84 & 81.32 & 90.32 & 28.22 & 56.19 & 70.19 & 32.36 & 64.96 & 81.01\\
 & 64x & 27.69 & 59.11 & 72.79 & 43.17 & 79.35 & 89.54 & 24.96 & 49.13 & 62.37 & 32.84 & 63.45 & 79.40  \\
 & 128x & 28.17 & 59.64 & 73.30 & 40.70 & 73.60 & 84.70 & 20.74 & 41.29 & 54.86 & 31.80 & 63.41 & 79.48  \\
 & 256x & 25.33 & 55.62 & 69.79 & 39.68 & 71.38 & 83.34 & 19.58 & 38.57 & 51.33 & 30.81 & 56.56 & 72.59  \\
 \cline{2-14}
\multirow{6}{*}{Sup-Phi4-mini} & 4x & 24.22 & 52.18 & 66.15 & 39.85 & 66.57 & 78.94 & 26.13 & 42.16 & 51.27 & 37.43 & 62.27 & 73.49  \\
 & 16x & 28.16 & 57.82 & 71.13 & 42.18 & 67.66 & 79.18 & 28.48 & 43.75 & 53.55 & 39.96 & 55.62 & 66.99  \\
 & 32x & 27.66 & 57.96 & 71.69 & 41.66 & 69.12 & 80.30 & 27.87 & 42.92 & 51.16 & 39.11 & 61.05 & 71.51  \\
 & 64x & 27.40 & 57.76 & 72.01 & 41.40 & 70.39 & 81.34 & 28.11 & 44.27 & 53.46 & 38.27 & 61.67 & 73.13   \\
 & 128x & 27.45 & 57.08 & 70.61 & 41.69 & 70.47 & 81.28 & 27.62 & 42.85 & 51.41 & 37.85 & 63.81 & 74.89 \\
 & 256x & 25.67 & 54.20 & 67.85 & 40.82 & 67.54 & 78.62 & 27.03 & 41.45 & 49.93 & 36.78 & 60.32 & 71.13 \\
\cline{2-14}
\multirow{6}{*}{CLaRa-Phi4-mini} & 4x & 8.58 & 28.21 & 44.61 & 18.38 & 42.90 & 60.01 & 10.56 & 26.81 & 40.08 & 17.05 & 38.03 & 52.63  \\
 & 16x & 20.94 & 48.89 & 64.34 & 23.90 & 49.17 & 64.38 & 15.07 & 30.79 & 42.49 & 21.16 & 46.83 & 64.10  \\
 & 32x & 27.63 & 58.37 & 72.37 & 28.64 & 55.48 & 70.00 & 16.43 & 34.31 & 46.55 & 30.30 & 58.89 & 75.98 \\
 & 64x & 29.17 & 60.08 & 73.83 & 36.25 & 65.67 & 78.44 & 16.38 & 33.36 & 46.23 & 27.74 & 53.69 & 69.63\\
 & 128x & 29.63 & 60.54 & 74.02 & 37.94 & 63.96 & 75.93 & 17.67 & 34.90 & 46.39 & 30.79 & 58.53 & 74.64 \\
 & 256x & 27.50 & 57.90 & 71.58 & 37.10 & 63.30 & 75.23 & 17.67 & 34.13 & 44.98 & 31.79 & 60.53 & 76.90\\

\hline
\rowcolor{gray!15}
\multicolumn{14}{c}{\textbf{Instruction-tuned-initialized}} \\
\multirow{6}{*}{Sup-Mistral-7B }& 4x & 28.33 & 58.52 & 71.96 & 42.40 & 65.93 & 77.40 & 27.87 & 45.47 & 55.39 & 39.90 & 62.02 & 74.35  \\
 & 16x & 28.20 & 57.24 & 69.57 & 42.24 & 67.91 & 79.33 & 26.80 & 42.33 & 50.68 & 39.22 & 59.77 & 70.38  \\
 & 32x & 27.56 & 56.70 & 69.58 & 44.88 & 71.02 & 81.54 & 29.29 & 43.86 & 52.22 & 42.21 & 55.47 & 66.56\\
 & 64x & 25.70 & 54.11 & 66.80 & 44.94 & 73.32 & 84.24 & 29.02 & 44.21 & 53.28 & 42.09 & 54.68 & 66.03\\
 & 128x & 25.05 & 53.12 & 65.73 & 45.14 & 74.56 & 85.18 & 28.11 & 43.18 & 52.40 & 42.27 & 58.47 & 69.94\\
 & 256x & 25.15 & 52.61 & 65.27 & 44.60 & 73.95 & 85.18 & 28.69 & 44.32 & 53.25 & 41.89 & 52.64 & 63.77\\
 \cline{2-14}

\multirow{6}{*}{CLaRa-Mistral-7B} & 4x & 24.66 & 55.27 & 69.82 & 25.73 & 52.20 & 68.63 & 18.06 & 37.67 & 50.27 & 28.51 & 57.74 & 73.00  \\
 & 16x & 23.66 & 52.43 & 66.79 & 35.85 & 67.69 & 81.13 & 18.95 & 37.21 & 50.71 & 12.85 & 32.17 & 47.90 \\
 & 32x & 21.54 & 49.50 & 65.13 & 35.78 & 65.94 & 80.11 & 16.75 & 36.93 & 50.48 & 17.14 & 39.59 & 55.75 \\
 & 64x & 20.77 & 49.52 & 64.71 & 33.65 & 63.93 & 78.47 & 16.13 & 34.13 & 46.41 & 17.40 & 38.10 & 51.61\\
 & 128x & 20.07 & 47.19 & 61.98 & 32.24 & 61.83 & 77.36 & 13.77 & 29.70 & 41.93 & 19.43 & 42.62 & 57.31 \\
 & 256x & 19.39 & 46.91 & 62.66 & 29.71 & 56.66 & 71.58 & 15.84 & 31.92 & 43.57 & 20.12 & 42.26 & 56.29  \\
 \cline{2-14}

\multirow{6}{*}{Sup-Phi4-mini} & 4x & 22.38 & 49.29 & 62.81 & 39.91 & 66.12 & 77.59 & 26.11 & 41.34 & 50.39 & 37.79 & 61.89 & 72.96 \\
 & 16x & 23.85 & 52.04 & 65.97 & 40.92 & 65.97 & 77.75 & 26.47 & 42.17 & 51.07 & 38.52 & 53.51 & 64.79 \\
 & 32x & 23.72 & 52.09 & 65.43 & 41.53 & 67.81 & 79.44 & 27.41 & 41.40 & 50.23 & 39.79 & 51.25 & 60.99 \\
 & 64x & 22.34 & 49.46 & 63.40 & 40.93 & 68.95 & 80.57 & 26.17 & 42.17 & 51.14 & 38.47 & 49.72 & 58.88 \\
 & 128x & 22.80 & 50.64 & 63.30 & 40.47 & 67.99 & 80.36 & 27.05 & 41.82 & 50.21 & 38.95 & 50.31 & 59.78 \\
 & 256x & 22.27 & 49.31 & 62.15 & 39.76 & 65.61 & 78.39 & 26.54 & 41.06 & 48.39 & 37.68 & 49.09 & 56.86 \\
\cline{2-14}
\multirow{6}{*}{CLaRa-Phi4-mini} & 4x & 3.98 & 17.02 & 31.76 & 13.19 & 33.04 & 48.40 & 7.60 & 21.09 & 31.03 & 20.40 & 39.25 & 52.00  \\
 & 16x & 8.15 & 25.82 & 40.84 & 18.00 & 41.31 & 57.10 & 7.82 & 18.24 & 28.19 & 21.16 & 40.80 & 51.96  \\
 & 32x & 16.94 & 42.64 & 58.30 & 31.30 & 58.02 & 71.88 & 13.71 & 29.99 & 40.74 & 29.85 & 53.24 & 66.04  \\
 & 64x & 17.63 & 44.29 & 59.33 & 32.09 & 59.98 & 74.35 & 15.93 & 32.42 & 44.22 & 29.77 & 56.59 & 71.59  \\
 & 128x & 21.90 & 50.34 & 65.14 & 30.53 & 55.15 & 68.30 & 12.81 & 27.30 & 37.66 & 25.54 & 47.53 & 61.28 \\
 & 256x & 19.32 & 45.89 & 60.39 & 31.18 & 56.05 & 69.80 & 12.24 & 26.19 & 37.40 & 27.54 & 51.71 & 64.10 \\

\hline
\end{tabular}
\end{table}

\begin{table}[H]
    \centering
        \caption{Pretraining data statistics for SCP. 
    The table reports the total number of training examples (Num.), average number of generated QA pairs or documents (Avg.pairs),  average input document length (Avg.inp) and  average generated text length (Avg.out) for Simple QA, Complex QA, and Paraphrased Documents..
}
    \label{tab:pretraining_data}

    \begin{tabular}{c|ccc}
    \bottomrule
         &Simple QA&Complex QA & Paraphrase Doc   \\
         \hline 
         Num. &2,000,000&2,000,000 &1,966,291 \\
         Avg.pairs& 7.80&4.62&1.00 \\
         Avg.inp &95.56&95.56&95.56 \\
         Avg.out &158.18&253.90&108.67 \\
    \toprule
    \end{tabular}
\end{table}
\begin{table}[H]
\caption{\centering Statistics of   experimental datasets.}
\centering
\footnotesize
\begin{tabular}{lcc}
\toprule
Datasets &
Training Data Size &
Evaluation Data Size \\
\midrule
Nature Question & 58,622 &  6,489 \\
HotpotQA & 90,185 &  7,384 \\
MusiQue  & 168,745 & 2,417  \\
2WikiMultiHopQA & 167,454  & 12,576 \\
\bottomrule
\end{tabular}
\label{tab:benchmarks}
\end{table}
\section{Detailed experimental setup}
\label{app:ex_setup}
\subsection{Datasets}

The \textit{pretraining} corpus consists of 2M documents and their corresponding 2M \textit{SimpleQA} sets, 
2M \textit{ComplexQA} sets, and 2M paraphrased documents. 
Detailed statistics on data composition and distribution are provided in Table~\ref{tab:pretraining_data}.

During the \textit{instruction tuning} stage of compression learning, we use question data from \textsc{COCOM} \citep{10.1145/3701551.3703527} , which contains 453k questions. We employ the \texttt{Mistral-7B} model and retrieve the top-5 most similar documents from the \texttt{Wikipedia-2021} corpus using dense retrieval. 
Given each query and its retrieved documents, the model is prompted to generate the corresponding answer, which serves as the gold target for instruction tuning.

For \textit{end-to-end} training, we use the training set of each benchmark individually, except for MuSiQue. 
Since MuSiQue is more challenging and difficult to converge when trained alone, we construct its training set by combining the training samples from HotpotQA, 2Wiki, and MuSiQue.
For each query, we first obtain its positive documents, and then retrieve additional documents from the corpus using the \texttt{BGE-large-en-v1.5} model until we collect a total of 20 candidates. 
This ensures that the gold answer remains inferable from at least one of the selected documents during end-to-end optimization. Table~\ref{tab:benchmarks} summarizes the data statistics.

\subsection{Models}
For document retrieval, we employ \texttt{BGE-large-en-v1.5}\footnote{https://huggingface.co/BAAI/bge-large-en-v1.5} as the retriever for coarse ranking. 
Unless otherwise specified, we adopt \texttt{Mistral-7B-Instruct-v0.2}\footnote{https://huggingface.co/mistralai/Mistral-7B-Instruct-v0.2} as the default backbone for all experiments. 
Additionally, we evaluate the proposed method on \texttt{
Phi-4-mini-instruct}\footnote{https://huggingface.co/microsoft/Phi-4-mini-instruct} to assess its generalization across different model families. 
On top of the backbone model, we implement three LoRA modules: a \textit{compressor}, a \textit{query reasoner}, and a \textit{generation module}.

\subsection{Evaluation Metrics}
\label{eva_metric}
Following previous studies \citep{10.5555/3737916.3741392,louis-etal-2025-pisco}, we evaluate the compressor using the \textbf{Cover Exact Match (ACC)} metric, which measures whether the ground-truth answer is included in the generated output. 
For the reranker, we report \textbf{Recall@k} ($k \in \{1,3,5\}$), defined as the proportion of positive documents appearing within the top-$k$ ranked results.
For the generation model, we adopt two standard QA metrics: \textbf{Exact Match (EM)} and \textbf{F1}. 
The EM score measures the percentage of predicted answers that exactly match the gold answers, while the F1 score computes the token-level overlap between predictions and references, reflecting the harmonic mean of precision and recall.

\begin{table}[H]
\centering
\small
\caption{\centering Hyperparameter settings used in our experiments.}
\label{tab:hyperparams}
\begin{tabular}{lc}
\toprule
\textbf{Hyperparameter} & \textbf{Value} \\
\midrule
LR Scheduler & cosine \\
Optimizer & AdamW \\
Epochs & 1 \\
LoRA Layers ($r$) & all-linear \\
LoRA Rank ($r$) & 16 \\
LoRA Dropout & 0.1 \\
LoRA Alpha & 32 \\
LoRA Rank ($r$) & 16 \\
Warmup Ratio & 0.03 \\
Max Gradient Norm & 1.0 \\
Documents Max Tokens & 256 \\
\rowcolor{gray!30} Compression learning & \\ 
$\lambda$  & 0.1 \\
Batch Size & 128 \\
Learning Rate (LR) & $1 \times 10^{-4}$ \\
\rowcolor{gray!30} End-to-end learning & \\ 
Batch Size & 32 \\
Learning Rate (LR) & $5 \times 10^{-6}$ \\
\bottomrule
\end{tabular}
\end{table}

\subsection{Implementation Details}
Table~\ref{tab:hyperparams} summarizes the hyperparameters used for all LoRA modules and training stages. 
Specifically, we employ separate configurations for the compression learning, and end-to-end training phases.
During end-to-end learning, both the query reasoner and the generator are initialized from the compressor-trained checkpoints. 
Following \cite{shi2025directretrievalaugmentedoptimizationsynergizing}, for each query $x$, we first retrieve the top-20 documents from the corpus using \texttt{BGE-large-en-v1.5}, obtain their corresponding compressed representations, and then pass them along with the query into the query reasoner to identify the top-$k$ ($k=5$) ranked documents, which are subsequently fed into the generator.

For corpus preprocessing, each document is segmented into chunks of 256 tokens. 
We extensively evaluate our model under different compression ratios 
$\rho \in \{4, 16, 32, 64, 128, 256\}$, 
where the number of memory tokens is computed as $256 / \rho$.
All experiments are conducted on $8\times100$ H100 GPUs. 
Unless otherwise stated, all training runs are performed for a single epoch.

\subsection{Baselines}
\label{app:baselines}
In this section, we provide detailed descriptions of all baseline methods used for comparison under different experimental settings. 
We categorize them into three groups: (1) \textit{compression baselines}, 
(2) \textit{retrieval and reranking baselines}, 
and (3) \textit{end-to-end QA baselines}.

\subsubsection{Compression Baselines}

\paragraph{AutoCompressor. \citep{chevalier-etal-2023-adapting}} 
This method segments a long document into chunks, appends a \texttt{<Sum>} token at the end of each chunk, 
and trains the model to produce a fixed number of summary vectors. 
During training, the model is fine-tuned with a standard language modeling cross-entropy loss, 
with a stop-gradient applied to past summary vectors. 
At inference time, the model first compresses and then reuses the summaries, 
achieving efficient long-context reasoning at significantly reduced cost.

\paragraph{XRAG. \citep{10.5555/3737916.3741392}}
XRAG treats retrieved document embeddings as an additional \textit{retrieval modality}, 
mapping them into the language model’s representation space via a lightweight projection layer. 
This enables retrieval-augmented generation with as few as a single ``document token.'' 
XRAG adopts a two-stage training strategy: 
(1) \textit{Paraphrase Pretraining} to align document embeddings with textual semantics, and 
(2) \textit{Context-Aware Instruction Tuning} with self-distillation to optimize retrieval utilization. 
Only the projection layer is trained, while both the retriever and language model remain frozen, 
achieving compression ratios up to 178×.

\paragraph{COCOM. \citep{10.1145/3701551.3703527}}
COCOM maps each retrieved document into a compact sequence of \textit{context embeddings} 
(e.g., compressing hundreds of tokens into 4–128 embeddings), 
which reduces input length and accelerates generation. 
It jointly trains a compressor and a generator with two objectives: 
(i) an auto-encoding reconstruction loss to preserve semantic information, and 
(ii) a conditional generation loss to ensure high-quality answers from compressed contexts. 
The framework also supports multi-document compression and cross-document fusion, 
and offers a lightweight variant (\textit{COCOM-light}) using BERT as the compressor.

\paragraph{PCC. \citep{dai-etal-2025-pretraining}}
PCC consists of an encoder and a transformer-based converter. 
The encoder extracts compact semantic representations, 
while the converter adjusts their dimensionality and semantics through two MLP layers 
so that the compressed memory can be directly fed into any LLM. 
The model is pretrained (with the LLM frozen) using a combination of 
\textit{auto-encoding reconstruction} and \textit{auto-regressive completion} tasks 
to retain generation-relevant information. 
Domain-specific fine-tuning is then performed on limited data for RAG QA, ICL reasoning, and dialogue tasks.

\paragraph{LLMLingua-2. \citep{pan-etal-2024-llmlingua}}
LLMLingua-2 constructs a large-scale extractive compression dataset using GPT-4-generated high-fidelity summaries. 
It formulates compression as a token-level binary classification problem (keep or remove), 
where a bidirectional Transformer encoder (e.g., XLM-RoBERTa) estimates the retention probability of each token. 
Tokens are ranked by their probabilities to achieve 2–5× compression 
while maintaining semantic completeness.

\paragraph{PISCO. \citep{louis-etal-2025-pisco}}
PISCO introduces trainable \textit{memory tokens} appended to the document, 
jointly fine-tuned with LoRA adapters to compress text by up to 1/16 of its original length. 
It employs \textit{sequence-level knowledge distillation (SKD)} from teacher-generated answer sequences 
to ensure consistency between compressed and uncompressed outputs.

\subsubsection{Retrieval and Reranking Baselines}

\paragraph{BM25.}
A classical lexical retrieval method that scores each document based on 
term frequency, inverse document frequency, and document length normalization.

\paragraph{BGE-Reranker. \citep{bge_m3}}
A recent large-scale, general-purpose reranker that directly predicts the relevance score 
between a query and each candidate document, used to reorder initial retrieval results.

\paragraph{RankZephyr. \citep{pradeep2023rankzephyreffectiverobustzeroshot} }
A 7B-parameter open-source reranker distilled in two stages from RankGPT-3.5 and RankGPT-4. 
It integrates variable-window training, input order shuffling, and teacher-guided ranking data, 
achieving robust performance under varying document counts and ranking conditions. 
During inference, RankZephyr performs iterative sliding-window ranking using prompt-decoder style generation.

\paragraph{Setwise. \citep{10.1145/3626772.3657813}}
Unlike pairwise reranking, Setwise compares multiple candidate documents in a single inference step, 
greatly reducing LLM calls and prompt length. 
It leverages classical sorting algorithms (e.g., heap or bubble sort) 
and directly estimates relevance probabilities from model logits, 
avoiding step-by-step list generation.

\paragraph{Rank-R1. \citep{zhuang2025rankr1enhancingreasoningllmbased}}
A reinforcement learning-based reranking framework 
that enhances LLM reasoning capabilities for document ranking. 
Built upon the Setwise ranking paradigm, 
it introduces explicit reasoning instructions before answer generation 
and optimizes the model via \textit{Group Relative Policy Optimization (GRPO)}. 
The model is trained only with queries and relevance labels, 
and receives reward signals based on prediction correctness and format compliance.

\subsubsection{End-to-End QA Baselines}

\paragraph{GenGround. \citep{shi-etal-2024-generate}}
This method decomposes complex questions into sub-questions 
using the model’s internal knowledge, 
then refines preliminary answers via retrieved documents for evidence grounding. 
It further introduces \textit{Instructional Grounding Distillation (IGD)}, 
which distills grounding trajectories from ChatGPT into smaller open models such as Mistral-7B.

\paragraph{In-Context RAG.}
Selects the top-$k$ retrieved documents using the BGE Reranker 
and feeds them as context to the LLM for direct answer generation.

\paragraph{ReComp. \citep{xu2024recomp}}
ReComp retrieves relevant documents and compresses them into concise, query-related summaries 
via either an extractive or a generative compressor. 
These summaries are then used as context for answer generation. 
Training jointly optimizes both retriever and compressor, 
allowing selective retrieval when documents are unhelpful.

\paragraph{DPA-RAG. \citep{dong2025understand}}
This method introduces preference-aligned retrieval and generation. 
It first constructs preference data by analyzing LLM responses under various retrievals 
and then aligns both reranker and generator 
through a hybrid of point-wise, pair-wise, and contrastive training objectives.

\paragraph{RetRobust. \citep{yoran2024making}}
Improves robustness of RAG systems through two mechanisms: 
(i) using an NLI model to filter irrelevant retrieved texts, 
and (ii) fine-tuning with mixed relevant/irrelevant retrieval samples 
so that the model learns when to utilize or ignore retrieval information.

\paragraph{ChatQA. \citep{10.5555/3737916.3738409}}
A context-augmented instruction-tuned model 
that integrates multi-source conversational and instruction data 
to enhance reasoning and refusal capabilities. 
It also fine-tunes a dense retriever on multi-turn QA data, 
replacing traditional query rewriting modules.

\paragraph{Self-RAG. \citep{asai2024selfrag}}
Incorporates \textit{reflection tokens} (e.g., ``need retrieval?'', ``retrieved relevant?'', ``supported answer?'') 
so the model can self-assess and adaptively decide when to retrieve external knowledge. 
Training combines GPT-4–generated annotated data with self-reflective labeling 
to enable dynamic retrieval and self-critique during inference.

\paragraph{RAG-DDR. \citep{li2025ragddr}}
Employs \textit{Differentiable Data Rewards (DDR)} 
to achieve fully end-to-end optimization of RAG systems. 
It uses rollout-based system rewards 
and aligns retrieval and generation through Direct Preference Optimization (DPO).

\paragraph{DRO. \citep{shi2025directretrievalaugmentedoptimizationsynergizing}}
Models document ordering as a latent variable 
and alternates between inference and optimization using a variational EM framework. 
The E-step estimates document order distributions via importance sampling, 
while the M-step jointly updates the selector and generator 
based on weighted likelihood maximization.

\section{Pretraining Data Quality}
\label{app:data-quality}
To ensure the quality of the constructed pretraining data, 
we conducted a manual evaluation. 
We randomly sampled 200 examples for each output type, 
resulting in a total of 600 samples, 
which were independently assessed by one of the authors. 
The evaluation results indicate that, 
thanks to our rigorous filtering process, 
almost all generated samples successfully cover the key information contained in the source documents. 
Only 21 instances exhibited mild hallucinations, 
where the model introduced information not present in the original text. 
This demonstrates that the synthesized data are of high factual and semantic quality, 
providing a reliable foundation for compression pretraining.
\begin{table}[H]
\scriptsize
\centering
\caption{Instruction tuning performance of \textsc{Mistral-7B} and \textsc{Phi4-mini} models 
under different pretraining corpus sizes (0.5M, 1M, 2M). 
Results are reported on four QA datasets under both \textit{Normal} and \textit{Oracle} retrieval settings 
with a fixed compression ratio (\textit{CR} = 32).}
\label{data_scaling}
\begin{tabular}{ccccccc}
\hline
Models & Corpus size & NQ & HotpotQA& Musique& 2Wikiqa & Average \\
\rowcolor{gray!15}
\multicolumn{7}{c}{\textbf{Normal}} \\
\multirow{3}{*}{Mistral-7B}  & 0.5M &53.38&41.40&10.30&46.67&37.94\\
  & 1M &54.82&43.71&10.63&46.90&39.02 \\
 & 2M &54.64&43.52&10.55&46.58 &38.82 \\

\hline 
\multirow{3}{*}{Phi4-mini} & 0.5M & 48.82&38.53&7.78&43.57 &34.67 \\
 & 1M & 48.40&38.47&7.73&43.82&34.61 \\
 & 2M & 49.30&38.62&7.70&43.71&34.83 \\

\hline
\rowcolor{gray!15}
\multicolumn{7}{c}{\textbf{Oracle}} \\
\multirow{3}{*}{Mistral-7B}  & 0.5M & 70.33&62.47&29.16& 57.97&54.98\\
& 1M & 74.08 &68.88&38.97&63.91 &61.46 \\
& 2M & 73.77 &69.51&38.31&64.54&61.53 \\

\hline 
\multirow{3}{*}{Phi4-mini} & 0.5M &68.31&64.41&29.25&58.22&55.05 \\
& 1M & 69.41&64.42&31.32&58.00&55.79 \\
 & 2M & 69.90&65.32&31.77&58.52&56.38 \\
\toprule
\end{tabular}
\end{table}

\section{Pretraining Data Scaling}
\label{app:data_scale}
To investigate how the number of pretraining samples affects the performance of the compressor, we train models with varying amounts of pretraining data and assess their performance after instruction tuning 
on four QA datasets. 
The results are illustrated in Table~\ref{data_scaling}. We observe that enlarging the pretraining corpus generally leads to consistent performance improvements across all datasets and both retrieval settings. 
For instance, under the \textit{Normal} setting, the Mistral-7B model improves its average score from 37.94 to 39.02 as the corpus size increases from 0.5M to 1M, while the performance remains stable when further scaled to 2M. 
A similar trend can be observed in the \textit{Oracle} setting, where the model achieves an average gain of over 6 points when moving from 0.5M to 1M, indicating that additional pretraining data enhances the compressor’s ability to preserve more task-relevant information.

For the smaller Phi4-mini model, the improvements are relatively modest, suggesting that model capacity may constrain the benefits of scaling pretraining data. 
Overall, these findings demonstrate that moderate expansion of pretraining data contributes positively to downstream QA performance, while extremely large pretraining sets bring diminishing returns.

\begin{figure}[h]
    \centering
    \includegraphics[width=0.8\linewidth]{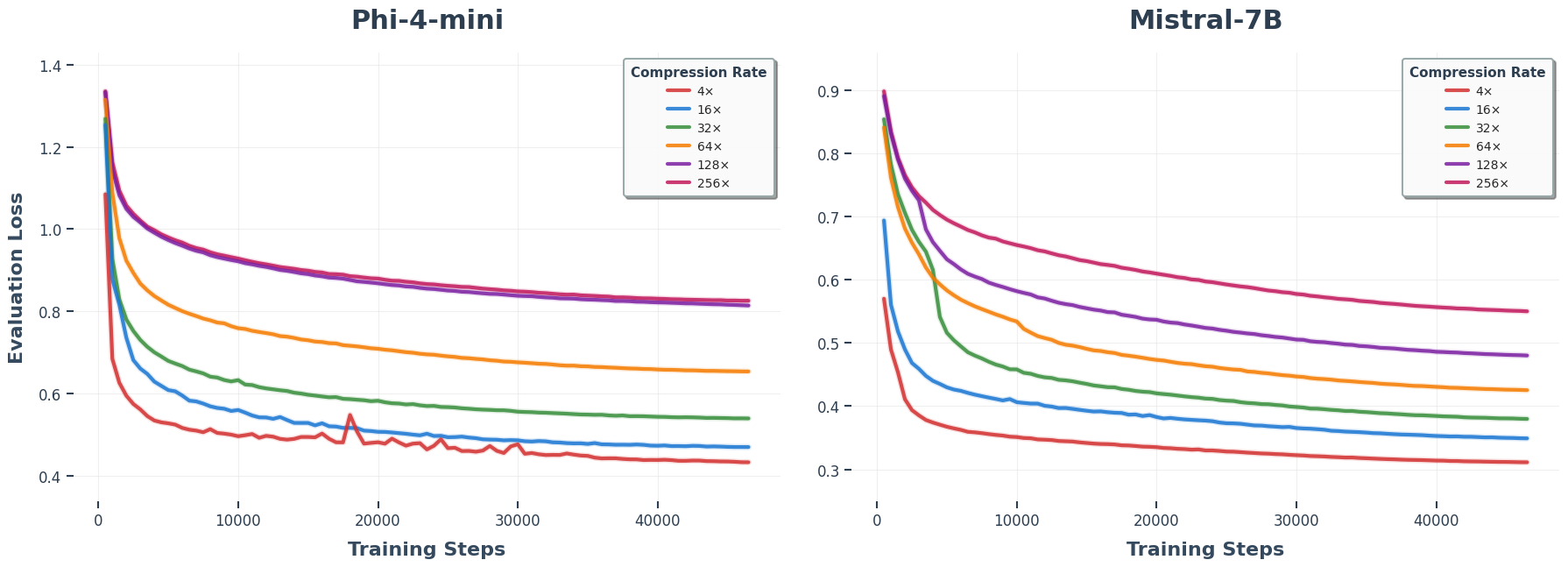}
    \caption{Validation loss curves during the compression pretraining stage under different compression ratios (CR) on the \texttt{Phi-4-mini} (left) and \texttt{Mistral-7B} (right) models. }
    \label{compression_stage1_curve}
\end{figure}

\begin{figure}[ht]
    \centering
    \includegraphics[width=1\linewidth]{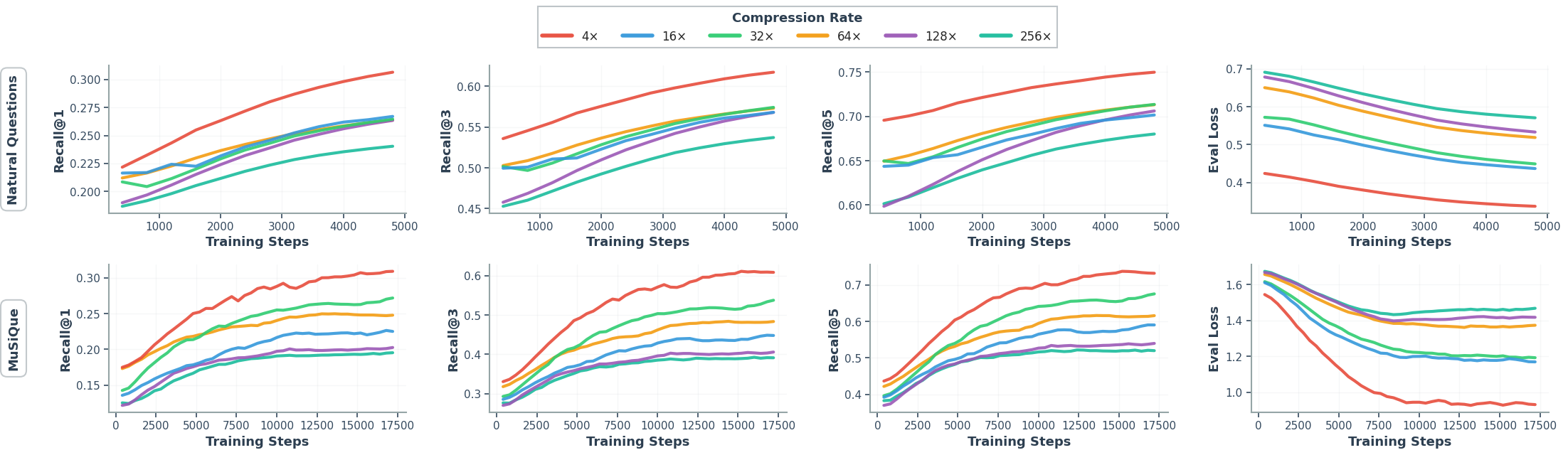}
    \caption{Validation trends of recall and evaluation loss during the end-to-end training stage under different compression ratios (CR) on the \textsc{NQ} (top) and \textsc{Musique} (bottom) datasets.}
    \label{end_to_end_curve}
\end{figure}

\section{Training Curves}
\label{app:train_curve}
Figures~\ref{compression_stage1_curve} present the validation loss curves during the compression pretraining stage across different compression ratios. 

A clear trend emerges: as the compression ratio increases, the validation loss rises for both models. 
This effect is more pronounced for \texttt{Phi4-mini}, where losses at ratios of 128 and 256 diverge sharply. 
In contrast, \texttt{Mistral-7B} exhibits relatively uniform loss gaps across compression ratios. 
We hypothesize that this difference arises because of capacity. \texttt{Phi4-mini}, with fewer parameters, has limited representational ability. At very high compression levels (e.g., CR=128), excessive information loss leads to semantic degradation and a steep rise in validation loss.

Fig~\ref{end_to_end_curve} presents the validation curves during end-to-end training on the \textsc{NQ} and \textsc{Musique} datasets. 
Recall scores consistently increase while evaluation losses steadily decrease, indicating stable and effective optimization. 
Higher compression ratios generally yield lower recall and higher loss, mirroring the trends observed during the compression pretraining stage.

\section{More Analysis}
\label{app:more_analysis}
\subsection{Effect of Freezing the Compressor and Query Reasoner}

We investigate the effect of limiting the fine-tuning scope to the generator module while freezing both the compressor and query reasoner. 
Specifically, we examine two representative compression settings, \textit{CR=32} and \textit{CR=128}, 
and compare model performance when only the generator is fine-tuned during both the instruction tuning and end-to-end QA training stages. 
The results are shown in Table~\ref{generator-only-instruction-tuning} and \ref{generator-only-end-to-end}. During the compression learning and instruction tuning stages, we observe that fine-tuning the compressor alongside the generator brings only marginal improvements. 
For example, under the \textit{Normal} setting, the average gain of full fine-tuning over generator-only tuning is less than 2.0\% across most datasets. 
Considering that in Section~\ref{sec_retrieval}, the instruction-tuned compressor tends to degrade retrieval performance due to its focus on answer-centric representations, 
a promising future direction is to explore how to effectively extract task-relevant information from compressed representations without directly fine-tuning the compressor itself. 
In contrast, during the end-to-end learning stage, fine-tuning the query reasoner proves to be more beneficial. 
A trainable retrieval module enables the model to identify more relevant documents and provide stronger contextual grounding for the generator. 
For instance, under the \textit{Oracle} setting with a compression ratio of 32, the F1 score of \textsc{Mistral-7B} improves from 52.54\% to 70.91\% when jointly fine-tuning both the query reasoner and generator. 
This highlights the crucial role of query reasoner in enhancing overall QA performance within our unified training framework.

\begin{table}[H]
\scriptsize
\centering
\caption{Instruction tuning results of \textsc{Mistral} and \textsc{Phi} models 
under different fine-tuning scopes (generator-only vs. finetune-both), 
retrieval modes (Normal vs. Oracle), and compression ratios (CR = 32, 128) 
on four QA datasets.}
\label{generator-only-instruction-tuning}
\begin{tabular}{ccccccc}
\hline
Models & CR & NQ & HotpotQA& Musique& 2Wiki & Average \\
\rowcolor{gray!15}
\multicolumn{7}{c}{\textbf{Normal}} \\
Generator-only \\
\hline 
Mistral&32x&52.26&42.66&10.43&45.79 &37.78\\
Mistral&128x&50.72&40.10&9.14&45.52 & 36.37\\
Phi&32x&45.91&37.70&6.95&42.97 &33.39 \\
Phi&128x&38.83&32.30&6.50&42.53 &30.04 \\
\hline 
Full finetune \\
\hline 
Mistral& 32x & 54.64 & 43.52 & 10.55 & 46.58 & 38.82 \\
Mistral& 128x & 53.36 & 41.37 & 10.26 & 46.40 & 37.85 \\
Phi&  32x & 49.30 & 38.62 & 7.70 & 43.71 & 34.83 \\
Phi&  128x & 43.09 & 33.92 & 6.87 & 43.70 & 31.90 \\
\hline 

\hline
\rowcolor{gray!15}
\multicolumn{7}{c}{\textbf{Oracle}} \\
Generator-only \\
\hline 

mistral&32x&72.78&67.48&34.38&60.89 &58.88\\ 
mistral&128x&66.93&59.66&25.94&58.19 & 52.68\\
phi&32x&65.65&62.76&27.60&56.46 & 53.12\\
phi&128x&52.87&47.51&17.38&48.98 & 41.68\\
\hline 
Full finetune \\
\hline 
mistral & 32x & 73.77 & 69.51 & 38.31 & 64.54 & 61.53 \\
mistral& 128x & 69.96 & 62.09 & 30.86 & 59.08 & 55.50 \\
phi& 32x & 69.90 & 65.32 & 31.77 & 58.52 & 56.38 \\
phi& 128x & 60.44 & 51.52 & 19.28 & 50.29 & 45.38 \\
\hline 
\end{tabular}
\end{table}

\begin{table}[H]
\scriptsize
\centering
    \caption{End-to-end QA performance of \textsc{Mistral} and \textsc{Phi} models 
under different fine-tuning scopes (generator-only vs. finetune-both), 
retrieval modes (Normal vs. Gold), and compression ratios (CR = 32, 128) 
on \textsc{HotpotQA} and \textsc{2WikiQA} datasets.
}
\label{generator-only-end-to-end}
\begin{tabular}{c|c|cc|cc}
\hline
\multirow{2}{*}{Models} & \multirow{2}{*}{CR} 
& \multicolumn{2}{c|}{HotpotQA}
& \multicolumn{2}{c}{2Wiki}
\\
\cline{3-6}
& &  F1 & EM
  & F1 & EM
 \\
\hline
\rowcolor{gray!15}
\multicolumn{6}{c}{\textbf{Normal}} \\
Generator-only \\
\hline 

Mistral  & 32x  &38.40&28.24&39.93&35.80\\
Mistral & 128x &38.26&28.26&41.34&37.29\\
Phi4  & 32x  &32.91&23.51&35.99&32.14\\
Phi4 & 128x &31.42&21.89&35.54&31.32\\
\hline 
Full finetune \\
\hline 
Mistral& 32x &    41.84 & 31.26 & 43.23 & 38.98  \\
Mistral& 128x & 42.26 & 31.78 & 41.80 & 37.37 \\
 Phi4& 32x &    37.14 & 26.99 & 38.15 & 33.82 \\
Phi4& 128x &   34.73 & 24.95 &36.41 & 32.23 \\

\hline 
\rowcolor{gray!15}
\multicolumn{6}{c}{\textbf{Oracle}} \\
Generator-only \\
\hline 

Mistral  & 32x  & 52.54&40.11&45.43&40.98\\
Mistral & 128x & 51.60&39.06&44.64&40.22\\

Phi4  & 32x &45.91&34.20& 40.33&36.12\\
Phi4 & 128x &39.46&28.06&37.05&32.57 \\
\hline 
Full finetune \\
\hline 
Mistral & 32x &    70.91 & 57.07 & 66.32 & 61.12 \\
 Mistral& 128x &   66.51 & 52.30 & 64.82 & 58.97 \\
Phi4 & 32x &    51.06 & 38.33 &  50.68 & 45.41  \\
Phi4 &  128x&   52.68 & 38.49 &  49.97 & 44.11  \\
\toprule
    \end{tabular}
    
\end{table}

\subsection{Retrieval number generalization}
\begin{figure}
    \centering
    \includegraphics[width=1\linewidth]{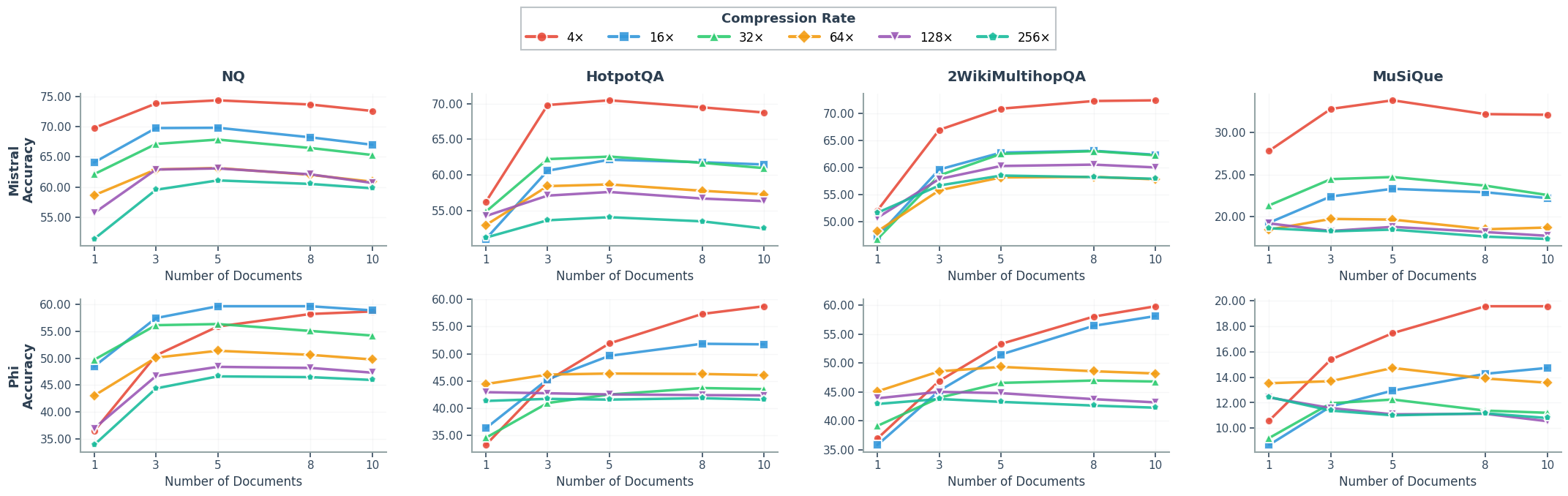}
    \caption{Performance of varying the number of retrieved documents ($k$) during testing on different QA datasets. 
}
    \label{diff_top_k}
\end{figure}

We further explore the impact of varying the number of retrieved documents (\textit{top-$k$}) during testing in our end-to-end training framework. 
During training, the model is consistently trained with the top-5 retrieved documents, 
while at test time, we vary $k$ from 1 to 10 to examine the model's sensitivity to retrieval size. 
The results are presented in Fig~\ref{diff_top_k}. 
As shown in the figure, the F1 score generally exhibits a rapid increase followed by a gradual decline as $k$ increases. 
However, the performance drop remains relatively small, indicating that our trained query reasoner and generator demonstrate good generalization 
with respect to the number of retrieved documents during inference.

\subsection{Effect of Query Reasoner Initialization}
We evaluate the effect of initializing the query reasoner with the pretrained compressor parameters 
versus random initialization on the \textsc{HotpotQA} and \textsc{2Wiki} datasets, as shown in Table~\ref{tab:reasoner_init}. 
The results demonstrate that compressor-initialized models consistently outperform their randomly initialized counterparts 
across all settings. 
This performance gain (e.g., from 66.84\%→70.91\% F1 and 62.68\%→66.32\% F1 on HotpotQA and 2Wiki, respectively) indicates that the pretrained compressor provides a strong prior for learning effective query reasoning representations, 
as it already encodes semantic relationships between queries and document content during the compression pretraining stage.

\begin{table}[H]
\scriptsize
\centering
\caption{End-to-End QA Performance with Randomly Initialized vs. Compressor-Initialized Query Reasoner}
\label{tab:reasoner_init}
\label{tab:qa_random_vs_copy}
\begin{tabular}{l|c|c|cc|cc}
\bottomrule
\multirow{2}{*}{\textbf{Model}} & \multirow{2}{*}{\textbf{CR}} & \multirow{2}{*}{\textbf{Retrieval Mode}} 
& \multicolumn{2}{c|}{\textbf{HotpotQA}} 
& \multicolumn{2}{c}{\textbf{2Wiki}} \\ 
\cline{4-7}
 &  &  & \textbf{F1} & \textbf{EM} & \textbf{F1} & \textbf{EM} \\ 
\hline
Mistral-7B & 32x & Normal & 39.48 & 29.12 & 39.90 & 35.79 \\
\quad w/ Compressor Init.
& 32x & Normal & 41.84 & 31.26 & 43.23 & 38.98 \\
Mistral-7B & 32x & Oracle & 66.84 & 52.91 & 62.68 & 57.55 \\
\quad w/ Compressor Init.
 & 32x & Oracle & 70.91 & 57.07 & 66.32 & 61.12 \\
Mistral-7B & 128x & Normal & 37.25 & 27.38 & 38.55 & 34.69 \\
\quad w/ Compressor Init.
& 128x & Normal & 42.26 & 31.78 & 41.80 & 37.37 \\
Mistral-7B & 128x & Oracle & 62.06 & 48.37 & 60.63 & 54.87 \\
\quad w/ Compressor Init.
& 128x & Oracle & 66.51 & 52.30 & 64.82 & 58.97 \\
\hline
\end{tabular}
\end{table}

\subsection{Efficiency Analysis}
We  evaluate the inference efficiency of our framework under different compression ratios. 
Specifically, for each query, we retrieve 20 candidate documents, compress them into 5 document representations using the compressor, 
and then generate the final answer based on these 5 compressed representations and the query. 
The average inference time for each stage is reported in Table~\ref{efficiency}. All timing statistics are measured on a single NVIDIA H100 GPU with 80GB memory.

As shown in the results, decoding with compressed representations takes only about 40\% of the time required when using full-text documents. 
Although compressing 20 documents is relatively time-consuming, this step can be performed offline; hence, it does not affect real-time inference latency during query answering. 
This makes the overall computational cost acceptable for practical deployment. 
We also observe that for the \textsc{Mistral} model, compression time tends to decrease as the compression ratio increases, 
while both decoding and query retrieval times remain relatively stable across different compression settings. 

\begin{table}[H]
\scriptsize
\centering
\caption{Average inference time (in milliseconds) for compression, retrieval, and decoding across different compression ratios (CR) on \textsc{Mistral-7B} and \textsc{Phi4-mini} models.}
\label{efficiency}
\begin{tabular}{c |c| c c c }
\bottomrule
\textbf{Models} & \textbf{CR} & \textbf{Compression Time} & \textbf{Query Time} & \textbf{Decoding Time}  \\
\hline  
Mistral-7B & Pure text& -- &--&1290.57\\
Mistral-7B& 4x& 1092.29&99.69&532.73\\
Mistral-7B& 16x& 922.85&94.17&502.78\\
Mistral-7B& 32x& 904.22&92.16&514.75 \\
Mistral-7B& 64x& 893.76&95.14&521.09\\
Mistral-7B& 128x&876.99&95.24&518.41 \\
Mistral-7B& 256x& 835.87&90.76&521.03 \\
\hline 
Phi4-mini&Pure text &--&--&870.29\\
Phi4-mini& 4x& 674.78&94.34&342.05\\
Phi4-mini& 16x& 574.46&89.53&343.01\\
Phi4-mini& 32x& 561.17&84.35&358.04\\
Phi4-mini& 64x&604.89&85.33&354.77 \\
Phi4-mini& 128x&594.55&91.73&360.23 \\
Phi4-mini& 256x&789.49&99.47&354.87 \\
\toprule
\end{tabular}
\end{table}

\section{Fidelity and Grounding Analysis}
\label{app:fied_ground}
In this section, we aim to understand how much essential information is retained in our compressed representations, 
and to what extent the generated answers remain grounded to the input documents and queries after both compression learning 
and end-to-end training.

\subsection{Information Preservation}

During compression representation pretraining, we include a paraphrasing objective that allows the generation model 
to reconstruct the original text from the compressed representation. 
We consider two evaluation settings: (1) \textit{unseen data}, consisting of positive documents of downstream QA tasks that were not used in pretraining, 
and (2) \textit{seen data}, where we randomly sample 4,000 documents from the pretraining corpus. 

We evaluate the reconstruction quality using several metrics: \textit{BERTScore} \citep{BERTScore} (which measures semantic similarity between texts), 
\textit{ROUGE-1} and \textit{ROUGE-L} (which capture lexical overlap), 
and following~\cite{improving_information}, we also compute the \textit{entity preservation ratio}, 
which measures the proportion of entities from the input text that are preserved in the reconstructed text\footnote{Entity extraction is performed using the SpaCy library.}.

The results are shown in Table~\ref{paraphrase_performance}. 
We observe that our model achieves a high BERTScore of nearly 90\%, which remains stable across different compression ratios. 
This indicates that the compressed representations successfully retain most of the semantic information from the original text. 
For ROUGE-1, ROUGE-L, and entity preservation, the model also maintains relatively high scores—over 50\% on average. 
We further observe that as the compression ratio increases, the lexical overlap and entity preservation metrics gradually decline, 
suggesting that fewer memory tokens make it harder to reconstruct the exact surface form of the original text. 
However, the consistently high semantic similarity scores imply that the key meaning is preserved. 
This phenomenon may indicate that when using fewer memory tokens, the model tends to generate paraphrased expressions 
to maintain the original semantics. 
We leave further exploration of this linguistic compression behavior for future work.
\begin{table}[H]
\footnotesize
\centering
\caption{Evaluation of information preservation under different compression ratios (CR) 
on seen and unseen documents using \textsc{BERTScore}, \textsc{ROUGE}, and entity preservation.
}
\label{paraphrase_performance}
\begin{tabular}{c|c|cccc|cccc}
\hline
\multirow{2}{*}{Models} & \multirow{2}{*}{CR}
& \multicolumn{4}{c|}{Seen Data}
& \multicolumn{4}{c}{Unseen Data} \\
\cline{3-10}
& & BERT & R-1 & R-L & Entity
  & BERT & R-1 & R-L & Entity \\
\hline
\multirow{6}{*}{Mistral-7B}& 4x & 90.67 & 55.88 & 40.12 & 54.78 & 91.45 & 59.74 & 44.09 & 60.04 \\
& 16x & 90.63 & 56.12 & 40.33 & 54.78 & 91.43 & 59.97 & 44.10 & 59.88 \\
& 32x & 90.56 & 56.21 & 40.10 & 53.91 & 91.39 & 60.28 & 43.98 & 59.33 \\
& 64x & 90.28 & 55.54 & 38.86 & 51.45 & 91.24 & 60.09 & 43.42 & 58.26 \\
& 128x & 89.84 & 54.12 & 36.56 & 47.75 & 91.00 & 59.61 & 42.48 & 55.75 \\
& 256x & 89.19 & 51.75 & 33.12 & 42.12 & 90.51 & 57.89 & 39.59 & 52.38 \\
\hline
\multirow{6}{*}{Phi4-mini} & 4x & 90.93 & 58.48 & 42.16 & 57.86 & 91.70 & 62.00 & 45.10 & 63.14 \\
 & 16x & 90.77 & 58.20 & 41.49 & 56.28 & 91.66 & 62.20 & 45.31 & 62.22 \\
 & 32x & 90.36 & 57.04 & 39.40 & 52.38 & 91.42 & 61.71 & 44.34 & 59.64 \\
 & 64x & 89.53 & 54.27 & 35.40 & 45.28 & 90.84 & 60.20 & 41.72 & 54.47 \\
 & 128x & 88.26 & 49.30 & 29.65 & 34.98 & 89.61 & 55.68 & 35.58 & 43.89 \\
 & 256x & 88.13 & 49.05 & 29.10 & 34.27 & 89.52 & 55.27 & 35.22 & 43.61 \\
\hline
\end{tabular}
\end{table}

\begin{table}[H]
\scriptsize
\centering
\caption{Grounding evaluation of \textsc{Mistral-7B} and \textsc{Phi4-mini} models 
under different initialization settings (pretraining-initialized vs. instruction-tuned-initialized), 
retrieval modes (Normal vs. Oracle), and compression ratios (CR). 
Metrics include \textit{Faithfulness (Faith)} and \textit{Factual Correctness (Fc)} across four QA datasets.
}
\label{grounding_performance}
\begin{tabular}{c|c|c|cc|cc|cc|cc|cc}
\hline
 \multirow{2}{*}{Models} & \multirow{2}{*}{CR} & \multirow{2}{*}{Retrieval}
& \multicolumn{2}{c|}{NQ}
& \multicolumn{2}{c|}{HotpotQA}
& \multicolumn{2}{c|}{Musique}
& \multicolumn{2}{c|}{2Wiki}
& \multicolumn{2}{c|}{Average} \\
\cline{4-13}
&  & & Faith & Fc
 & Faith & Fc
 & Faith & Fc
 & Faith & Fc
 & Faith & Fc \\
\hline
\rowcolor{gray!15}
\multicolumn{13}{c}{\textbf{Pretraining-initialized}} \\
\multirow{12}{*}{ Mistral-7B }& 4x & \multirow{6}{*}{ Normal} & 81.57 & 8.39 & 67.42 & 11.80 & 55.80 & 9.57 & 56.45 & 5.15 & 65.31 & 8.73 \\
 & 16x & & 75.85 & 8.97 & 62.50 & 13.58 & 49.64 & 10.34 & 51.02 & 5.04 & 59.75 & 9.48 \\
  & 32x &  & 72.65 & 7.34 & 61.40 & 11.86 & 51.35 & 8.18 & 53.30 & 7.30 & 59.67 & 8.67 \\
  & 64x &  & 67.49 & 8.29 & 57.95 & 10.85 & 46.55 & 10.00 & 44.44 & 5.16 & 54.11 & 8.57 \\
  & 128x &  & 65.99 & 8.67 & 56.50 & 10.35 & 44.52 & 9.50 & 43.26 & 4.54 & 52.57 & 8.27 \\
  & 256x &  & 64.74 & 7.42 & 53.68 & 12.34 & 40.79 & 7.87 & 40.88 & 4.85 & 50.02 & 8.12 \\
 \cline{2-13}
        &4x  & \multirow{6}{*}{Oracle} & 86.73 & 9.42 & 83.75 & 18.72 & 67.67 & 13.60 & 80.16 & 6.26 & 79.58 & 12.00 \\

   & 16x &  & 81.13 & 9.74 & 83.16 & 19.19 & 63.89 & 11.05 & 73.96 & 4.12 & 75.54 & 11.03 \\

   & 32x &  & 79.34 & 8.16 & 81.87 & 16.31 & 65.27 & 10.79 & 71.60 & 3.71 & 74.52 & 9.74 \\

   & 64x &  & 74.43 & 8.07 & 78.39 & 16.63 & 55.89 & 10.09 & 67.61 & 4.30 & 69.08 & 9.77 \\

   & 128x &  & 74.63 & 10.40 & 77.17 & 12.60 & 55.14 & 10.47 & 61.62 & 4.00 & 67.14 & 9.37 \\

   & 256x &  & 71.13 & 9.63 & 73.69 & 16.49 & 51.35 & 10.22 & 56.03 & 4.67 & 63.05 & 10.25 \\
\hline
 \multirow{12}{*}{Phi4-mini} & 4x &  \multirow{6}{*}{Normal}   & 79.45 & 7.47 & 62.46 & 10.06 & 50.57 & 11.13 & 51.05 & 5.27 & 54.69 & 8.48 \\
  & 16x &  & 74.99 & 8.13 & 61.45 & 12.84 & 51.61 & 10.04 & 48.59 & 5.17 & 59.16 & 9.05 \\
  & 32x &  & 66.58 & 5.74 & 59.22 & 11.64 & 45.44 & 9.03 & 45.16 & 3.69 & 54.10 & 7.52 \\
  & 64x &  & 59.66 & 8.36 & 50.44 & 9.42 & 39.52 & 8.53 & 39.79 & 4.13 & 47.35 & 7.61 \\
  & 128x &  & 60.26 & 8.18 & 51.83 & 10.57 & 35.08 & 9.47 & 34.72 & 4.61 & 45.48 & 8.20 \\
  & 256x &  & 57.65 & 7.78 & 45.57 & 8.71 & 33.79 & 7.38 & 32.73 & 5.71 & 42.44 & 7.39 \\
 \cline{2-13}
        &  4x&  \multirow{6}{*}{Oracle}  & 82.51 & 8.32 & 82.88 & 18.79 & 60.67 & 11.02 & 76.81 & 4.19 & 75.72 & 10.58 \\

   &  16x&  & 81.00 & 8.81 & 81.73 & 17.55 & 63.38 & 10.54 & 69.61 & 3.03 & 73.93 & 9.98 \\

   & 32x &  & 75.42 & 10.20 & 74.79 & 16.72 & 54.49 & 10.30 & 61.67 & 3.26 & 66.59 & 10.12 \\

   & 64x &  & 68.72 & 7.66 & 71.63 & 15.60 & 49.55 & 7.51 & 56.36 & 2.63 & 61.56 & 8.35 \\

   &  128x&  & 63.49 & 8.99 & 66.46 & 14.84 & 44.17 & 8.38 & 48.50 & 2.07 & 55.65 & 8.57 \\

  & 256x &  & 61.36 & 9.80 & 64.48 & 11.60 & 44.01 & 10.76 & 53.23 & 3.10 & 55.77 & 8.82 \\
\hline
\rowcolor{gray!15}
\multicolumn{13}{c}{\textbf{Instruction-tuned-initialized}} \\
 \multirow{12}{*}{Mistral-7B} & 4x & \multirow{6}{*}{Normal} & 54.67 & 39.69 & 30.21 & 36.15 & 12.68 & 11.08 & 5.50 & 38.97 & 25.76 & 31.47 \\
  & 16x &  & 52.13 & 39.13 & 37.68 & 39.91 & 13.12 & 12.55 & 11.13 & 36.17 & 28.52 & 31.94 \\
  & 32x &  & 49.33 & 37.64 & 36.08 & 36.97 & 12.25 & 10.29 & 12.93 & 40.87 & 27.65 & 31.44 \\
  & 64x &  & 50.60 & 36.92 & 35.35 & 36.71 & 11.84 & 10.52 & 13.50 & 34.24 & 27.82 & 29.60 \\
  & 128x &  & 49.03 & 36.40 & 36.52 & 39.58 & 11.77 & 10.56 & 12.13 & 35.57 & 27.36 & 30.53 \\
  & 256x &  & 50.90 & 40.02 & 34.95 & 38.75 & 11.81 & 11.41 & 15.73 & 37.67 & 28.35 & 31.96 \\
\cline{2-13} 
        & 4x & \multirow{6}{*}{Oracle} & 69.55 & 71.58 & 67.94 & 68.23 & 23.86 & 37.92 & 38.43 & 64.61 & 49.95 & 60.59 \\

   & 16x &  & 69.57 & 67.79 & 60.25 & 60.83 & 18.85 & 27.60 & 39.90 & 60.19 & 47.14 & 54.10 \\

  & 32x &  & 65.97 & 63.60 & 59.77 & 61.72 & 21.91 & 30.38 & 33.40 & 56.57 & 45.26 & 53.07 \\

   & 64x &  & 63.90 & 62.81 & 57.47 & 56.46 & 18.60 & 21.99 & 27.40 & 52.27 & 41.84 & 48.38 \\

   & 128x &  & 68.57 & 60.98 & 55.75 & 57.86 & 16.19 & 21.43 & 29.40 & 55.38 & 42.48 & 48.91 \\

   & 256x &  & 65.60 & 63.81 & 55.46 & 56.30 & 16.80 & 20.74 & 28.40 & 52.97 & 41.57 & 48.46 \\
\hline 
 \multirow{12}{*}{Phi4-mini} & 4x & \multirow{6}{*}{Normal}  & 49.50 & 41.02 & 28.13 & 35.57 & 12.65 & 13.56 & 6.37 & 34.37 & 24.16 & 31.13 \\
  & 16x &  & 41.57 & 30.75 & 27.70 & 32.59 & 9.82 & 8.21 & 7.87 & 35.37 & 21.74 & 26.73 \\
  & 32x &  & 41.07 & 26.85 & 28.64 & 30.10 & 8.58 & 8.97 & 10.63 & 31.36 & 22.23 & 24.32 \\

  & 64x &  & 35.98 & 25.25 & 28.57 & 29.71 & 11.37 & 11.03 & 10.13 & 34.87 & 21.51 & 25.21 \\

  & 128x &  & 38.47 & 26.75 & 30.23 & 29.52 & 8.06 & 8.25 & 11.03 & 31.00 & 21.95 & 23.88 \\

  & 256x &  & 37.50 & 27.89 & 26.48 & 25.10 & 9.34 & 9.03 & 10.80 & 31.74 & 21.03 & 23.44 \\
\cline{2-13}
       & 4x & \multirow{6}{*}{Oracle}  & 65.13 & 59.84 & 43.67 & 52.32 & 14.57 & 22.90 & 18.37 & 48.38 & 35.43 & 45.86 \\

   & 16x &  & 66.45 & 62.90 & 46.57 & 52.31 & 13.28 & 18.58 & 21.87 & 47.04 & 37.04 & 45.21 \\

   & 32x &  & 60.20 & 53.13 & 46.27 & 45.89 & 13.70 & 17.42 & 25.30 & 43.71 & 36.37 & 40.04 \\

   & 64x &  & 57.35 & 52.66 & 47.40 & 48.79 & 13.18 & 18.28 & 18.63 & 43.99 & 41.13 & 40.93 \\

   & 128x &  & 55.30 & 51.05 & 42.34 & 42.43 & 12.80 & 17.51 & 18.67 & 44.84 & 32.28 & 38.96 \\

   & 256x &  & 52.80 & 47.28 & 45.35 & 46.32 & 11.98 & 14.93 & 23.50 & 40.64 & 33.41 & 37.29 \\
\hline
\end{tabular}
\end{table}
\subsection{Grounding Analysis}

We further evaluate the grounding quality between the generated answers and the compressed document representations 
under both compression evaluation and end-to-end evaluation settings. 
We adopt the \textsc{RAGAs}~\citep{es2025ragasautomatedevaluationretrieval} package, which implements the \textit{LLM-as-a-Judge} paradigm for assessing generation quality. 
Two key metrics are used: \textit{faithfulness}, which measures whether the generated answer is faithful to the provided context 
and relevant to the query, and \textit{factual correctness}, which evaluates whether the answer is factually supported by the context. 
We employ GPT-4o-mini as the judging model.

The results are presented in Table~\ref{grounding_performance}. 
For the compression evaluation, our model achieves consistently high faithfulness scores, particularly when positive documents are included, 
indicating that the model generates answers more closely aligned with the query. 
However, the factual correctness scores are comparatively lower, consistent with findings reported in~\citet{improving_information}. 
We hypothesize that this is because, after instruction tuning, the generation model tends to produce longer and more elaborative answers, 
occasionally introducing tokens that do not appear in the original context. 
We also observe a decreasing trend in faithfulness as the compression ratio increases and model size decreases. 

For the end-to-end evaluation, the model demonstrates strong performance across both metrics. 
In particular, under the Mistral-7B model with a compression ratio of 4 and the top-20 retrieval setting containing positive documents, 
faithfulness and factual correctness reach 49.95 and 60.59, respectively. 
The higher factual correctness is likely due to the use of short gold answers during training, 
which encourages the model to directly copy relevant words or phrases from the retrieved documents. 
Other observed trends are consistent with those in the compression evaluation results.

\section{Paraphrase Case Study}
To better understand the nature of information captured by our compressed representations and the query reasoner, we analyze how effectively the compressed representations preserve key semantic information. 
As illustrated in Table~\ref{tab:paraphrase_doc_tcolor}, 
the generations conditioned on compressed representations largely preserve the essential semantic content of the original documents, 
while substantially altering surface expressions such as sentence structure and word order. 
This observation is consistent with our pretraining objective, 
which encourages the model to encode core semantic meaning rather than memorize lexical sequences. 
The model’s ability to reconstruct paraphrased variants of the original text demonstrates that 
the learned compressed representations successfully capture high-level semantic knowledge in a compact form.

\begin{table}[!ht]
  \centering
  \caption{Examples of Original and Paraphrased Documents generated from compressed representations.}
\label{tab:paraphrase_doc_tcolor}
\begin{tcolorbox}[
    colback=gray!3,      
    colframe=gray!40,   
    arc=2mm,             
    boxrule=0.4pt,       
    left=3mm, right=3mm, top=2mm, bottom=2mm,
    width=\linewidth,
    title=\textbf{Examples of Original and Paraphrased Documents Generated from Compressed Representations},
    coltitle=black,
    fonttitle=\bfseries,
]
\small

\textbf{Example 1} \\[2pt]
\textbf{Original Document:}  
This a document about SMK Bukit Kepayang. 1) Tuan Haji Mat Rohani Bin Kamar was the first Principal of the School. (2004 - 2008); 2) Madam Hajah Sadariah Binti Haji Md. Jali is the current Principal of the school. (2008-2014). Madam Hajah Sadariah achieved the title Master Principle in the year 2011 due to outstanding performance and achievements in that year.\\[3pt]

\textbf{Paraphrased Document:}  
From 2004 to 2008, \textit{Tuan Haji Mat Rohani Bin Kamar} served as the Principal of SMK Bukit Kepayang. 
He was succeeded by \textit{Madam Hajah Sadariah Binti Haji Md. Jali}, who led the school until 2014. 
In 2011, she was honored with the title of \textit{Master Principal} for her exceptional achievements. \\[3pt]

\rule{\linewidth}{0.3pt}

\textbf{Example 2} \\[2pt]
\textbf{Original Document:}  
This is a document about Sikorsky SH-60 Seahawk and the U.S. 5th Fleet. 
The ALMDS is a sensor system designed to detect, classify, and localize floating and near-surface moored mines  
in littoral zones, straits, and choke points. The system is operated from an MH-60S, which gives it a countermine role traditionally handled by the MH-53E Sea Dragon, allowing smaller ships the MH-53E can't operate from to be used in the role. The ALMDS beams a laser into the water to pick up reflections, then produces a video image for ground personnel to determine if the object is a mine. The MH-60S will utilize the BAE Systems Archerfish.\\[3pt]

\textbf{Paraphrased Document:}  
The ALMDS system, designed to detect and classify floating and near-surface mines in strategic locations such as straits, littoral zones, and choke points, is operated from an MH-60S helicopter. This platform enables the U.S. Navy’s 5th Fleet to perform a countermine mission that was previously the responsibility of the MH-53E Sea Dragon. The MH-60S is equipped with the BAE Systems Archerfish system, which plays a key role in the process. The ALMDS emits a laser pulse into the water, which reflects.  \\[3pt]

\rule{\linewidth}{0.3pt}

\textbf{Example 3} \\[2pt]
\textbf{Original Document:}  
This a document about Pinewild Women's Championship. The Pinewild Women's Championship was a golf tournament on the LPGA Tour, played only in 1995. It was played at the Pinewild Country Club of Pinehurst in Pinehurst, North Carolina. Rosie Jones was the winner, beating Dottie Pepper on the first hole of a sudden-death playoff. \\[3pt]

\textbf{Paraphrased Document:}  
In 1995, the Pinewild Women's Championship took place as a single-year event on the LPGA Tour. The competition was held at the Pinewild Country Club of Pinehurst, located in Pinehurst, North Carolina. Rosie Jones emerged victorious, securing the title by defeating Dottie Pepper in a sudden-death playoff on the first hole. \\[3pt]

\end{tcolorbox}
\end{table}

\begin{figure}[htbp]
    \centering
    \begin{subfigure}{0.4\textwidth}
        \centering
        \includegraphics[width=\linewidth]{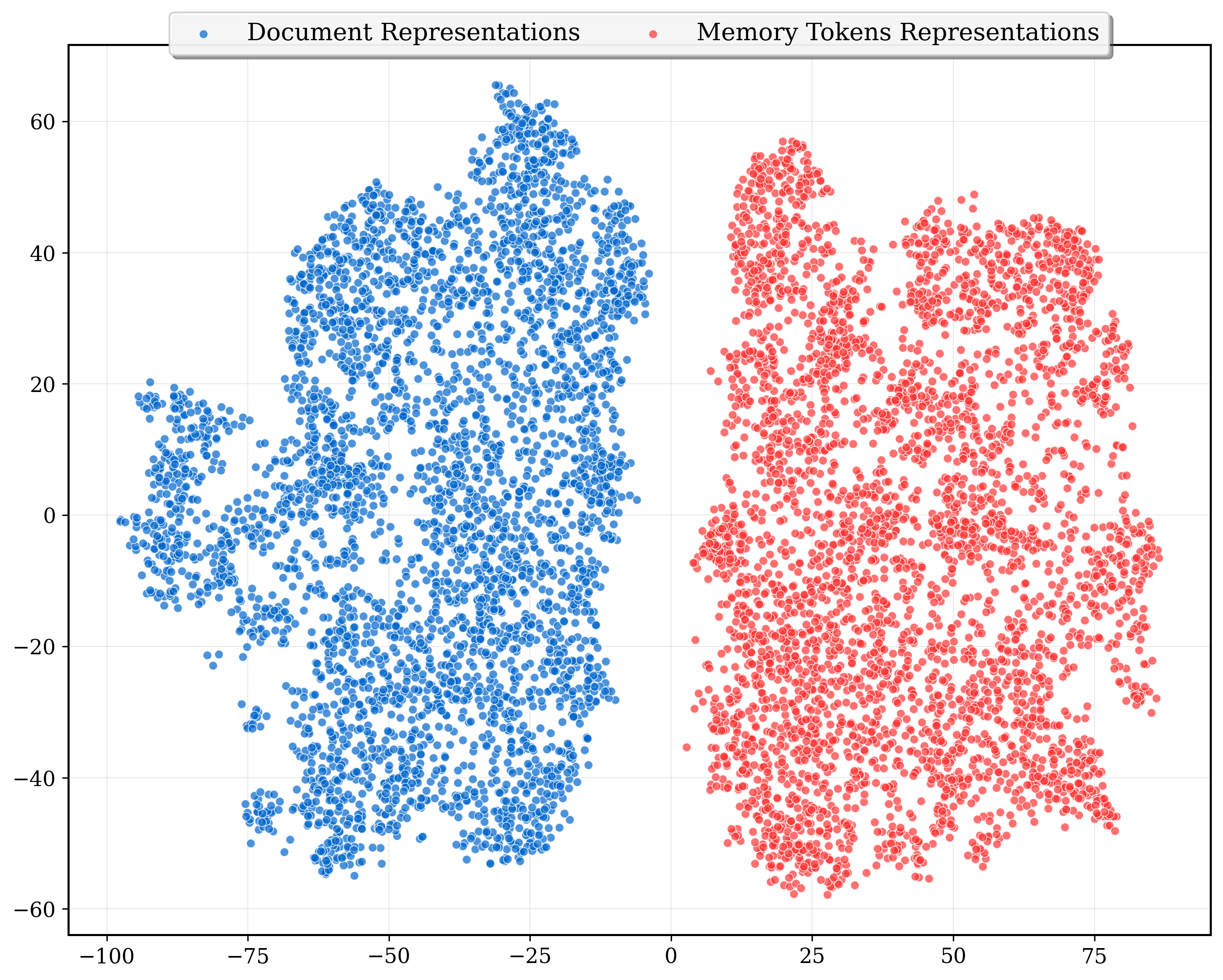}
        \caption{Without MSE}
        \label{fig:no_mse_side}
    \end{subfigure}
    \hfill
    \begin{subfigure}{0.4\textwidth}
        \centering
        \includegraphics[width=\linewidth]{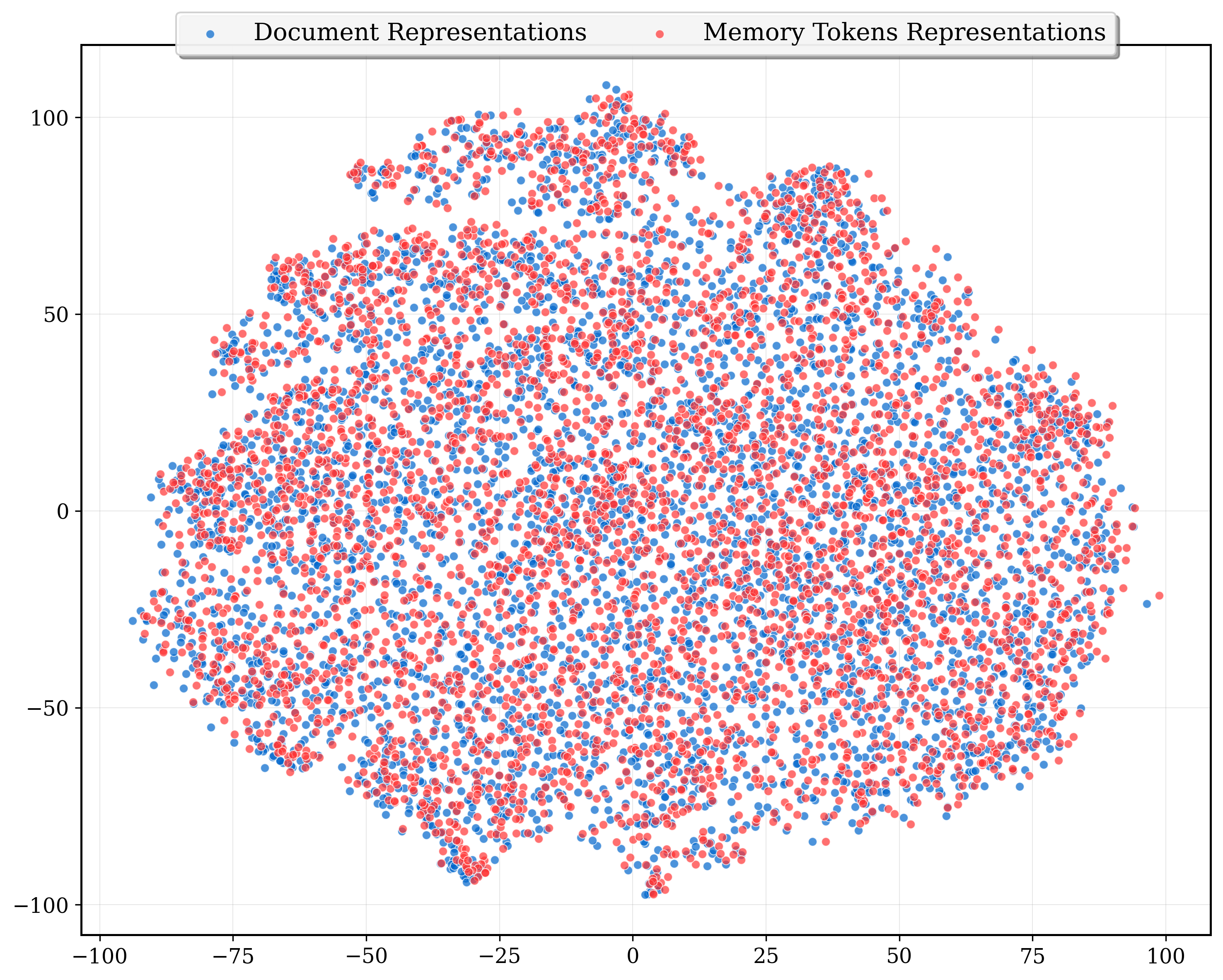}
        \caption{With MSE}
        \label{fig:with_mse_side}
    \end{subfigure}
    \caption{t-SNE visualization of document representations and compressed (memory token) representations of \textsc{Mistral-7B} under the compression ratio of 32. }
    \label{fig:mse_loss}
\end{figure}

\section{Prompts}
\label{appendix:prompt}
Figures~\ref{prompt_factual_question_generation}--\ref{prompt_paraphrase_validation} 
illustrate the prompts used during the data synthesis process. 
Specifically, we employ different prompting strategies for 
(1) generating QA pairs, 
(2) producing paraphrased documents, 
(3) validating information completeness, 
and (4) completing missing information. 
Additionally, Fig~\ref{prompt_doc_based_qa} 
shows the prompt template used by the generation model 
to answer questions based on the compressed document representations.

\begin{figure*}[!thbp]
\centering
\begin{minipage}{0.95\textwidth}
\begin{promptbox}[\color{black}{Prompt for Simple Question Generation}]{lightblue}{prompt:factual-question-generation}
\label{temp:factual_question_generation_prompt}

\ttfamily
You are given a document delimited by <doc> and </doc>. Your job is to read the given document and generate a comprehensive set of multi-hop questions that fully cover all the key information in the text. \\[4pt]

<doc> \\
<INSERT DOCUMENT HERE> \\
</doc> \\[4pt]

Question Requirements: \\[2pt]
You should generate as many questions as necessary to fully cover all the key facts in the document. \\
(1) Each question must be self-contained, meaning it should be understood by the user without seeing the document. \\
(2) Each question must cover only one or at most two distinct key pieces of information. \\
(3) The questions must be non-overlapping — no two questions should target the same piece of information. \\
(4) The questions should be simple factual recall only — do not require inference, reasoning, or summarization. \\
(5) Your output should be a list of self-contained, non-overlapping factual questions that together comprehensively cover all the key information in the document. \\[6pt]

There are some examples: \\
\{3 demonstrations\} \\[6pt]

Your output should be a JSON object with the following format: \\
\{ \\
\ \ \ "Question1": "...", \\
\ \ \ "Question2": "...", \\
\ \ \ ..., \\
\ \ \ "QuestionN": "..." \\
\} 
\end{promptbox}
\end{minipage}
\caption{Prompt used for simple question generation.}
\label{prompt_factual_question_generation}
\end{figure*}

\begin{figure*}[t]
\centering
\begin{minipage}{0.95\textwidth}
\begin{promptbox}[\color{black}{Prompt for Complex Question Generation}]{lightblue}{prompt:question-generation}
\label{temp:question_generation_prompt}

\ttfamily

You are given a document delimited by <doc> and </doc>. Your job is to generate a set of MULTI-HOP questions that, taken together, comprehensively cover the document’s key information. \\
<doc> \\
<INSERT DOCUMENT HERE> \\
</doc> \\[4pt]

Question Requirements: \\
1) Self-contained: Every question must be understandable without viewing the document. \\
2) Multi-hop only: Each question must require at least TWO independent pieces of evidence from DIFFERENT parts of the document (e.g., different paragraphs/sections/tables/items). If a question can be answered from a single sentence or data point, REJECT it. \\
3) Non-overlapping: No two questions may target the same fact or the same combination of facts. Each question must have a unique reasoning path and evidence combination. \\
4) Coverage: Produce as many questions as needed to cover ALL key facts in the document. Prefer many small, precise multi-hop questions over a few large ones. \\
5) Focus: Each question should target ONE multi-hop objective, typically integrating 2–3 facts (bridging, comparison, aggregation, temporal/causal linking, entity–attribute joining, etc.). Do NOT bundle multiple unrelated sub-questions. \\
6) Verifiability: The answer to each question must be derivable SOLELY from the document, with no external knowledge or subjective judgment. \\
7) Clarity: Avoid yes/no questions and vague wording. Use explicit constraints, quantities, and identifiers where relevant. \\
8) No explanations: Do NOT include rationales, steps, or references—ONLY output the questions as JSON. \\
9) You can generate 2-hops, 3-hops, 4-hops, etc. questions. \\[6pt]

QUESTION TEMPLATES (use as patterns, adapt as needed) \\
- Bridging: "Which X satisfies BOTH condition A mentioned in [context A] AND condition B mentioned in [context B]?" \\
- Comparison: "Considering [pivot], which of X or Y meets [criterion] when combining details from [source 1] and [source 2]?" \\
- Aggregation: "When combining [quantity/info] from section A with [quantity/info] from section B, which single entity matches [combined constraint]?" \\
- Temporal/Causal: "Based on the timeline described in parts A and B, which event/entity fulfills [temporal/causal relation]?" \\[6pt]

There are some examples: \{3 demonstrations\} \\[6pt]

Input FORMAT: \\
Document: \\
<INSERT DOCUMENT HERE> \\[6pt]

OUTPUT FORMAT \\
Return ONLY a JSON object with keys "Question1", "Question2", …, "QuestionN". \\
Example (structure only): \\
\{  "Question1": "...", \\
\ \ \ "Question2": "...", \\
\ \ \ "Question3": "...", \\
\ \ \ "QuestionN": "..." 
\} \\
\end{promptbox}
\end{minipage}
\caption{Prompt used for complex question generation.}
\label{prompt_question_generation}
\end{figure*}

\begin{figure*}[t]
\centering
\begin{minipage}{0.95\textwidth}
\begin{promptbox}[\color{black}{Prompt for Answer Generation}]{lightblue}{prompt:answer-generation}
\label{temp:answer_generation_prompt}

\ttfamily
You are a factual answering assistant. \\
\\
Your task is to read the provided document and answer the given question **based only on the information explicitly stated in the document**. \\
Please output the answer as short as possible. \\
Requirements: \\
- Your answer must be based solely on the content of the document. \\
- Do not use prior knowledge or make assumptions beyond the document. \\
- If the document does not contain the answer, respond with: "The document does not contain this information." \\
- The answer should be concise, factual, and complete. \\[6pt]

Input Format: \\
Document: \\
<INSERT DOCUMENT TEXT HERE> \\
\\
Question: \\
<INSERT QUESTION HERE> \\[6pt]

Output Format: \\
Answer: <YOUR ANSWER HERE> 
\end{promptbox}
\end{minipage}
\caption{Prompt used for factual answer generation.}
\label{prompt_answer_generation}
\end{figure*}

\begin{figure*}[t]
\centering
\begin{minipage}{0.95\textwidth}
\begin{promptbox}[\color{black}{Prompt for QA Validation }]{lightblue}{prompt:qa-validation}
\label{temp:qa_validation_prompt}

\ttfamily
You are a fact-checking assistant. \\
\\
Your task is to verify whether the given answer to a question is **fully supported by the provided document**. \\
\\
Instructions: \\
- Read the document carefully. \\
- Read the question and the provided answer. \\
- Determine whether the answer is correct **based solely on the information in the document**. \\
- The answer must be **complete**, **factually correct**, and **not contain any information that is not in the document**. \\
\\
If the answer is fully correct and supported by the document, respond with: \\
"Correct" \\
\\
If the answer is partially correct, incomplete, or includes unsupported information, respond with: \\
"Incorrect" \\
\\
Input Format: \\
Document: \\
<INSERT DOCUMENT HERE> \\
\\
Question: \\
<INSERT QUESTION HERE> \\
\\
Answer: \\
<INSERT ANSWER HERE> \\
\\
Output Format: \\
\{\{"Judgment": "Correct" / "Incorrect"\}\} 
\end{promptbox}
\end{minipage}
\caption{Prompt used for QA validation.}
\label{prompt_qa_validation}
\end{figure*}


\begin{figure*}[t]
\centering
\begin{minipage}{0.95\textwidth}
\begin{promptbox}[\color{black}{Prompt for Supplementary Simple QA Generation}]{lightblue}{prompt:supplementary-factual-qa}
\label{temp:supplementary_factual_qa_generation_prompt}

\ttfamily
You are given a document and a set of existing question-answer pairs. Your task is to carefully compare the information covered in the QA pairs against the document and generate additional questions that cover any key information not yet addressed. \\[4pt]

Requirements: \\
- Only generate questions for key facts present in the document that are **not already covered** in the existing QA pairs. \\
- Do **not** repeat or rephrase the information in the existing question answer pairs. \\
- Each question should cover **only one or two distinct key pieces of information**. \\
- Each question must be self-contained, meaning it should be understood by the user without seeing the document. \\
- All questions should require **simple factual recall only**, with no inference or reasoning. \\[6pt]

There are some examples: \\
\{3 demonstrations\} \\[6pt]

Input Format: \\
Document: \\
<INSERT DOCUMENT HERE> \\
Existing QA: \\
<INSERT EXISTING QA HERE> \\[6pt]

Output Format: \\
Return your generated new supplementary questions in the following JSON format: \\
\{ \\
\ \ \ "Number of Supplementary Questions": N, \\
\ \ \ "Question1": "...", \\
\ \ \ "Question2": "...", \\
\ \ \ ..., \\
\ \ \ "QuestionN": "..." \\
\} \\[6pt]

If all key information is already covered and no supplementary questions are needed, output an empty JSON object: \\
\{ \\
\ \ \ "Number of Supplementary Questions": 0 \\
\} 
\end{promptbox}
\end{minipage}
\caption{Prompt used for supplementary Simple QA generation.}
\label{prompt_supplementary_factual_qa_generation}
\end{figure*}

\begin{figure*}[t]
\centering
\begin{minipage}{1\textwidth}
\begin{promptbox}[\color{black}{Prompt for Supplementary Complex QA Generation}]{lightblue}{prompt:supplementary-qa-generation}
\label{temp:supplementary_qa_generation_prompt}
\small
\ttfamily
You are given a document and a set of existing question-answer pairs. Your task is to carefully compare the information covered in the QA pairs against the document and generate additional MULTI-HOP questions that cover any key information not yet addressed. \\
Requirements: \\
- Only generate questions for key facts present in the document that are **not already covered** in the existing question answer pairs. \\
- Do **not** repeat or rephrase information which can be found in the existing question answer pairs. \\
Question Requirements: \\
1) Self-contained: Every question must be understandable without viewing the document. \\
2) Multi-hop only: Each question must require at least TWO independent pieces of evidence from DIFFERENT parts of the document (e.g., different paragraphs/sections/tables/items). If a question can be answered from a single sentence or data point, REJECT it. \\
3) Non-overlapping: No two questions may target the same fact or the same combination of facts. Each question must have a unique reasoning path and evidence combination. \\
4) Coverage: Produce as many questions as needed to cover ALL key facts in the document. Prefer many small, precise multi-hop questions over a few large ones. \\
5) Focus: Each question should target ONE multi-hop objective, typically integrating 2–3 facts (bridging, comparison, aggregation, temporal/causal linking, entity–attribute joining, etc.). Do NOT bundle multiple unrelated sub-questions. \\
6) Verifiability: The answer to each question must be derivable SOLELY from the document, with no external knowledge or subjective judgment. \\
7) Clarity: Avoid yes/no questions and vague wording. Use explicit constraints, quantities, and identifiers where relevant. \\
8) No explanations: Do NOT include rationales, steps, or references—ONLY output the questions as JSON. \\
9) You can generate 2-hops, 3-hops, 4-hops, etc. questions. \\[6pt]
QUESTION TEMPLATES (use as patterns, adapt as needed) \\
- Bridging: "Which X satisfies BOTH condition A mentioned in [context A] AND condition B mentioned in [context B]?" \\
- Comparison: "Considering [pivot], which of X or Y meets [criterion] when combining details from [source 1] and [source 2]?" \\
- Aggregation: "When combining [quantity/info] from section A with [quantity/info] from section B, which single entity matches [combined constraint]?" \\
- Temporal/Causal: "Based on the timeline described in parts A and B, which event/entity fulfills [temporal/causal relation]?" \\[6pt]
Input Format: \\
Document: <INSERT DOCUMENT HERE>  Existing QA: <INSERT EXISTING QA HERE> \\[6pt]
Output Format: \\
Note, do not repeat or paraphrase existing questions. Instead, generate new multi-hop questions for the missing information, and put the new questions in JSON format: \\
\{ \\
\ \ \ "Number of Supplementary Questions": N, \\
\ \ \ "Question1": "...", \\
\ \ \ ..., \\
\ \ \ "QuestionN": "..." \\
\} \\
\\
If all key information is already covered and no supplementary questions are needed, output an empty JSON object: \\
\{ \\
\ \ \ "Number of Supplementary Questions": 0 \\
\}
\end{promptbox}
\end{minipage}
\caption{Prompt used for supplementary complex QA generation.}
\label{prompt_supplementary_qa_generation}
\end{figure*}

\begin{figure*}[t]
\centering
\begin{minipage}{0.95\textwidth}
\begin{promptbox}[\color{black}{Prompt for Document Paraphrasing}]{lightblue}{prompt:paraphrase-generation}
\label{temp:paraphrase_generation_prompt}

\ttfamily
You are given a document. Your task is to paraphrase the document in a way that: \\[4pt]

(1) **Restructure extensively** — **Try your best to break down the structure of the original document**, do not keep the same paragraphing, ordering, or sentence flow. Reorganize ideas, shuffle the order, merge or split sentences, and restructure arguments. \\[4pt]

(2) **Preserve meaning with absolute accuracy** — ensure that ALL key information and semantics of the original document are retained. Do not omit any factual details, numbers, dates, or specific information. \\[4pt]

(3) **Avoid direct copying** — no sentence should remain identical; re-express ideas in a fresh way using synonyms and varied sentence structures. \\[4pt]

(4) **CRITICAL: Add no new information** — the paraphrased document cannot introduce: \\
\ \ \ - New facts, interpretations, or context not explicitly stated \\
\ \ \ - Organizational names or affiliations not mentioned in the original \\
\ \ \ - Explanatory details or background information \\
\ \ \ - Your own analysis or conclusions about the content \\[4pt]

(5) **Maintain the original's voice and perspective** — if the original uses commands ("followers shall..."), don't change it to descriptions ("the organization promotes..."). Preserve the document's intended tone and format. \\[4pt]

(6) **Verify factual relationships** — when paraphrasing complex information involving multiple entities, dates, or cause-and-effect relationships, double-check that you maintain the correct connections and chronology. \\[4pt]

(7) **Use varied vocabulary** — employ synonyms and alternative expressions while maintaining precision. \\[4pt]

(8) **Preserve completeness** — if the original mentions specific numbers, dates, names, or measurements, include them in your paraphrase (even if reworded). \\[4pt]

(9) **Maintain coherence** — the paraphrased version should read as natural and fluent writing. \\[4pt]

**WARNING: Do not assume context.** If a document mentions "the Samaj" without identifying what organization this refers to, do not assume it's "Brahma Samaj" or any other specific group. Work only with what is explicitly stated. \\[4pt]

Produce a paraphrased version that keeps the meaning and all factual details but has significantly altered structure and wording. \\[4pt]

Here are some examples of paraphrased documents. \\[2pt]
\{10 demonstrations\} \\[4pt]
Input: \\[2pt]
<document>
\end{promptbox}
\end{minipage}
\caption{Prompt used for document paraphrasing.}
\label{prompt_paraphrase_generation}
\end{figure*}

\begin{figure*}[t]
\centering
\begin{minipage}{0.95\textwidth}
\begin{promptbox}[\color{black}{Prompt for Paraphrase Validation}]{lightblue}{prompt:paraphrase-validation}
\label{temp:paraphrase_validation_prompt}

\ttfamily
You are given two texts: \\
Original Document – the source text containing the key information. \\
Paraphrased Document – a rewritten version of the original. \\[4pt]

Your task is to check whether the paraphrased document fully preserves all the key information from the original document, without adding any new information. \\[4pt]

Guidelines: \\[2pt]
(1) ``Key information'' means the essential facts, arguments, data, and main ideas of the original. \\
(2) The paraphrased document must: \\
(2.1) Contain all the key information from the original. \\
(2.2) Not omit any important point. \\
(2.3) Not introduce information that is absent in the original. \\
(2.4) Preserve meanings without distortion or contradiction. \\
(2.5) Differences in style, sentence structure, or wording are acceptable as long as the meaning is preserved. \\[4pt]

(3) Output format: \\
(3.1) Answer ``Yes'' if the paraphrased document completely retains all key information and introduces no new information. \\
(3.2) Answer ``No'' if any key information is lost, altered, or if extra/unwarranted information appears. \\[4pt]

Don't output any explanation. \\[4pt]

Input: \\[2pt]
Original Document: [insert original doc] \\
Paraphrased Document: [insert paraphrased doc] \\[6pt]

Output: \\[2pt]
Yes / No 
\end{promptbox}
\end{minipage}
\caption{Prompt used for paraphrase validation.}
\label{prompt_paraphrase_validation}
\end{figure*}

\begin{figure*}[t]
\centering
\begin{minipage}{0.95\textwidth}
\begin{promptbox}[\color{black}{Prompt for Document-based QA}]{lightblue}{prompt:doc-based-qa}
\label{temp:doc_based_qa_prompt}

\ttfamily
You are a helpful assistant. Your task is to extract relevant information from provided documents and to answer to questions as briefly as possible.

Background: \{docs\}

Question:\{question\}
\end{promptbox}
\end{minipage}
\caption{Prompt used for document-based question answering.}
\label{prompt_doc_based_qa}
\end{figure*}

\begin{table}[H]
\centering
\scriptsize
\caption{Effect of pretraining data composition on instruction-tuning performance
under Normal (top-5 retrieval) settings under the 32 compression ratio. We report the absolute score change (±) for each pretraining data setting relative to the No-pretrain baseline.}
\label{data_composition1}
\begin{tabular}{ccccccc}
\hline
Models & Data composition & NQ & HotpotQA& Musique& 2Wikiqa & Average \\
\hline
\rowcolor{gray!15}
\multicolumn{7}{c}{\textbf{Normal}} \\
\multirow{5}{*}{Mistral-7B} & No-pretrain & 53.03 & 40.63 & 9.68 & 46.64 & 37.50 \\
 & SimpleQA & 53.84{\tiny\textcolor{teal}{+0.81}} & 42.20{\tiny\textcolor{teal}{+1.57}} & 10.26{\tiny\textcolor{teal}{+0.58}} & 46.68{\tiny\textcolor{teal}{+0.04}} & 38.25{\tiny\textcolor{teal}{+0.75}} \\
 & Para & 54.52{\tiny\textcolor{teal}{+1.49}} & 43.05{\tiny\textcolor{teal}{+2.42}} & 10.51{\tiny\textcolor{teal}{+0.83}} & 46.41{\tiny\textcolor{red}{-0.23}} & 38.62{\tiny\textcolor{teal}{+1.12}} \\
 & SimpleQA+ComplexQA & 55.48{\tiny\textcolor{teal}{+2.45}} & 43.00{\tiny\textcolor{teal}{+2.37}} & 10.67{\tiny\textcolor{teal}{+0.99}} & 46.39{\tiny\textcolor{red}{-0.25}} & 38.88{\tiny\textcolor{teal}{+1.38}} \\
 & SimpleQA+ComplexQA+Para & 54.64{\tiny\textcolor{teal}{+1.61}} & 43.52{\tiny\textcolor{teal}{+2.89}} & 10.55{\tiny\textcolor{teal}{+0.87}} & 46.58{\tiny\textcolor{red}{-0.06}} & 38.82{\tiny\textcolor{teal}{+1.32}} \\
\hline
\multirow{5}{*}{Phi4-mini} & No-pretrain & 48.10 & 37.65 & 7.61 & 44.68 & 34.51 \\
 & SimpleQA & 48.56{\tiny\textcolor{teal}{+0.46}} & 38.91{\tiny\textcolor{teal}{+1.26}} & 8.19{\tiny\textcolor{teal}{+0.58}} & 43.70{\tiny\textcolor{red}{-0.98}} & 34.84{\tiny\textcolor{teal}{+0.33}} \\
 & Para & 48.65{\tiny\textcolor{teal}{+0.55}} & 38.41{\tiny\textcolor{teal}{+0.76}} & 7.74{\tiny\textcolor{teal}{+0.13}} & 44.11{\tiny\textcolor{red}{-0.57}} & 34.73{\tiny\textcolor{teal}{+0.22}} \\
 & SimpleQA+ComplexQA & 49.47{\tiny\textcolor{teal}{+1.37}} & 38.88{\tiny\textcolor{teal}{+1.23}} & 8.03{\tiny\textcolor{teal}{+0.42}} & 43.96{\tiny\textcolor{red}{-0.72}} & 35.08{\tiny\textcolor{teal}{+0.57}} \\
 & SimpleQA+ComplexQA+Para & 49.30{\tiny\textcolor{teal}{+1.20}} & 38.62{\tiny\textcolor{teal}{+0.97}} & 7.70{\tiny\textcolor{teal}{+0.09}} & 43.71{\tiny\textcolor{red}{-0.97}} & 34.83{\tiny\textcolor{teal}{+0.32}} \\
\toprule
\end{tabular}
\end{table}

\begin{table}[H]
\centering
\scriptsize
\caption{Instruction-tuning performance with and without MSE loss under different compression ratios (CR = 32, 128) and normal retrieval settings.}
\label{tab:mse_loss1}
\begin{tabular}{cclllll}
\bottomrule
Models & CR & NQ & HotpotQA& Musique& 2Wikiqa & Average \\
\rowcolor{gray!15}
\multicolumn{7}{c}{\textbf{Normal}} \\
Mistral-7B & 32x & 54.25 & 43.11 & 9.85 & 45.84 & 38.26 \\
\qquad w/ mse & 32x & 54.64{\tiny\textcolor{teal}{+0.39}} & 43.52{\tiny\textcolor{teal}{+0.41}} & 10.55{\tiny\textcolor{teal}{+0.70}} & 46.58{\tiny\textcolor{teal}{+0.74}} & 38.82{\tiny\textcolor{teal}{+0.56}} \\
Mistral-7B & 128x & 52.98 & 41.32 & 10.22 & 46.23 & 37.69 \\
\qquad w/ mse & 128x & 53.36{\tiny\textcolor{teal}{+0.38}} & 41.37{\tiny\textcolor{teal}{+0.05}} & 10.26{\tiny\textcolor{teal}{+0.04}} & 46.40{\tiny\textcolor{teal}{+0.17}} & 37.85{\tiny\textcolor{teal}{+0.16}} \\
\toprule
\end{tabular}
\end{table}

\end{document}

%% file: math_commands.tex
\usepackage{amsmath,amsfonts,bm}

\def\eqref#1{equation~\ref{#1}}

\def\1{\bm{1}}

\DeclareMathAlphabet{\mathsfit}{\encodingdefault}{\sfdefault}{m}{sl}
\SetMathAlphabet{\mathsfit}{bold}{\encodingdefault}{\sfdefault}{bx}{n}

